\documentclass[10pt,twocolumn,letterpaper]{article}

\usepackage{cvpr}              % To produce the CAMERA-READY version
\usepackage[dvipsnames]{xcolor}

\usepackage{multirow}
\usepackage{subcaption}
\usepackage[textsize=scriptsize]{todonotes}

\definecolor{cvprblue}{rgb}{0.21,0.49,0.74}
\usepackage[pagebackref,breaklinks,colorlinks,citecolor=cvprblue]{hyperref}
\usepackage{stix}
\usepackage{xcolor,colortbl}
\usepackage{xfrac}
\usepackage{float}

\newcommand{\afk}[1]{{#1}}

\def\paperID{} % *** Enter the Paper ID here
\def\confName{CVPR}
\def\confYear{2024}

\title{Layout-to-Image Generation with Localized Descriptions using ControlNet with Cross-Attention Control.}

\author{Denis Lukovnikov \qquad Asja Fischer\\
Ruhr University Bochum\\
{\tt\small \{denis.lukovnikov, asja.fischer\}@rub.de}}

\begin{document}
% \showthe\linewidth
% \showthe\textwidth
% \showthe\columnwidth

\maketitle

\begin{abstract}

While text-to-image diffusion models can generate high-quality images from textual descriptions, they generally lack fine-grained control over the visual composition of the generated images.
Some recent works tackle this problem by training the model to condition the generation process on additional input describing the desired image layout.
Arguably the most popular among such methods, ControlNet, enables a high degree of control over the generated image using various types of conditioning inputs (e.g. segmentation maps). 
However, it still lacks the ability to take into account localized textual descriptions that indicate which image region is described by which phrase in the prompt.

In this work, we show the limitations of ControlNet for the layout-to-image task and enable it to use localized descriptions using a training-free approach that modifies the cross-attention scores during generation.
We adapt and investigate several existing cross-attention control methods in the context of ControlNet and identify shortcomings that cause failure (concept bleeding) or image degradation under specific conditions.
To address these shortcomings, we develop a novel cross-attention manipulation method %that modifies attention weights less drastically 
in order to maintain image quality while improving control.
Qualitative and quantitative experimental studies focusing on challenging cases are presented, demonstrating the effectiveness of the investigated general approach, and showing the improvements obtained by the proposed cross-attention control method.

\end{abstract}

\begin{figure}[t]
    \centering
    \begin{minipage}{0.28\textwidth}
        \centering
        \begin{subfigure}{0.49\textwidth}
        \centering
            \includegraphics[width=0.75\textwidth]{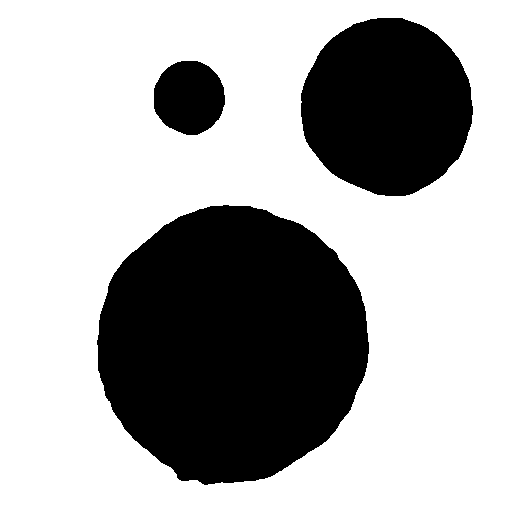}
            \caption{\label{fig:example:layers:1}``wooden table''}
        \end{subfigure}
        \hfill
        \begin{subfigure}{0.49\textwidth}
        \centering
            \includegraphics[width=0.75\textwidth]{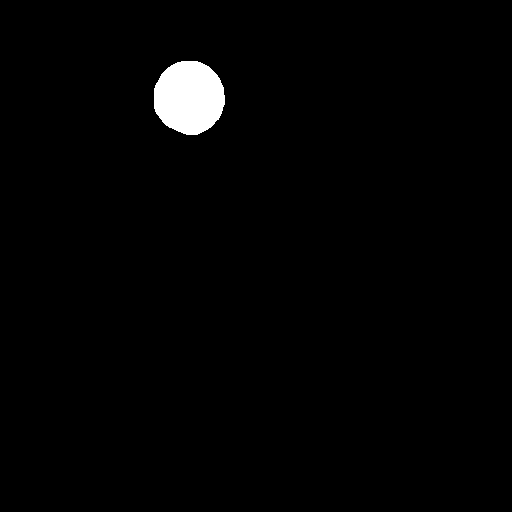}
            \caption{\label{fig:example:layers:2}``gold coin''}
        \end{subfigure}
        
        \medskip
        
        \begin{subfigure}{0.49\textwidth}
        \centering
            \includegraphics[width=0.75\textwidth]{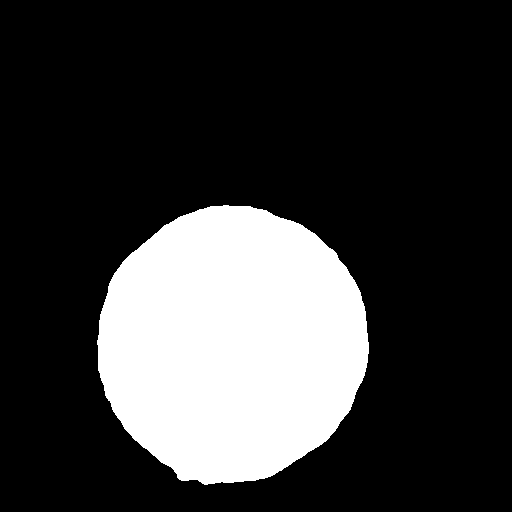}
            \caption{\label{fig:example:layers:3}``blue crystal ball''}
        \end{subfigure}
        \hfill
        \begin{subfigure}{0.49\textwidth}
        \centering
            \includegraphics[width=0.75\textwidth]{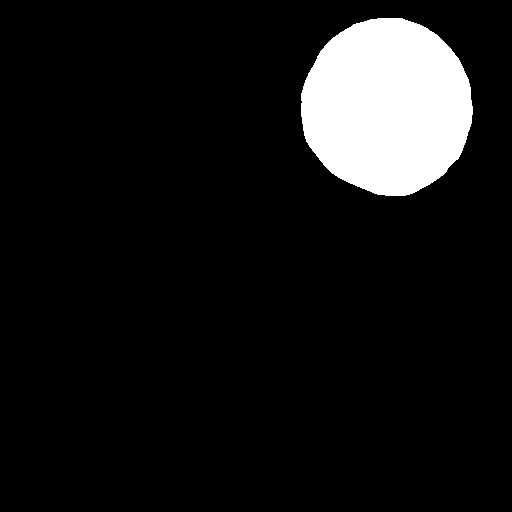}
            \caption{\label{fig:example:layers:4}``red tennis ball''}
        \end{subfigure}
    \end{minipage}%
    \hfill
    \begin{minipage}{0.19\textwidth}
        \centering
        \begin{subfigure}{\textwidth}
        \centering
            \includegraphics[width=0.75\textwidth]{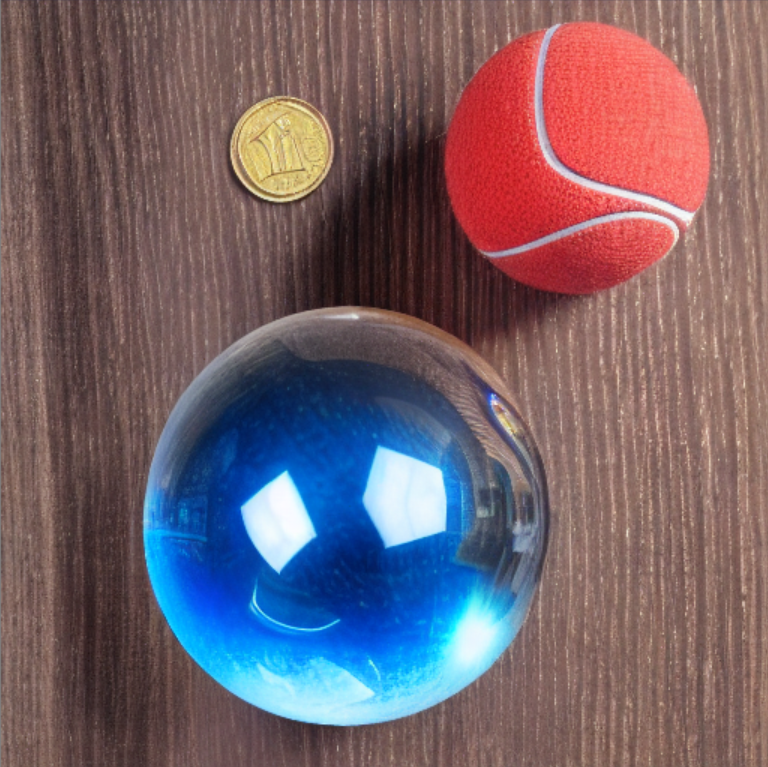}
            \caption{
            \label{fig:example:output}Desired output image for prompt ``\textit{\textcolor{black}{A photo of a \{\textcolor{blue}{blue crystal ball}\}$_c$, a \{\textcolor{red}{red tennis ball}\}$_d$ and a \{\textcolor{orange}{gold coin}\}$_b$ on a \{\textcolor{brown}{wooden table}\}$_a$}}''}
        \end{subfigure}
    \end{minipage}%
    
    \caption{An example of the task. The input consists of masks (a)-(d) and the annotated prompt in the caption of (e). The desired output is shown in (e). See \cref{sec:task}.}
\end{figure}

%%%%%%%%% BODY TEXT
\section{Introduction}
\label{sec:intro}
Diffusion-based text-to-image models like StableDiffusion~\cite{stablediffusion,kandinsky,dalle,imagen,ddpm,ddim} can generate high-quality images of various types of subjects from textual description. 
However, they lack fine-grained control over the visual composition of the generated image, which would make this a more useful tool for visual artists.
The default training method does not address generation scenario's where additional control inputs can be used to describe the desired composition of the image (e.g., using line art or segmentation maps).

Recent work has explored fine-tuning adapters (e.g. ControlNet~\cite{controlnet}, GLIGEN~\cite{gligen}, T2I-Adapters~\cite{t2iadapters}) to the diffusion model's U-Net in order to incorporate such signals, providing a high degree of control over the structure and content of the image.
However, such techniques generally still leave a lot of freedom to the model regarding object assignment, especially for non-trivial compositions.
For example, ControlNet lacks the ability to use localized descriptions %see Section~\ref{sec:task} and Fig.~\ref{fig:example:output}) 
in order to control what objects should be generated in each region.
% While ControlNet allows us to control the image layout using segmentation maps, %\footnote{Other types of control signals are also supported by ControlNet but do not fit our use case.}, 
% it lacks a mechanism to control what is generated inside each region.

The contributions of this work are two-fold.
Firstly, we investigate several training-free attention-based extensions of ControlNet to improve its grounding with a given localized textual description and identify important %aspects
\afk{characteristics} of such methods.
Secondly, we develop a novel cross-attention (CA) control method that avoids shortcomings of previous methods. 
A qualitative study shows superior performance with highly challenging inputs, confirmed by an automated evaluation on the newly developed \textsc{SimpleScenes} dataset. 
An evaluation using COCO2017 indicates that image quality remains on the same level as with plain ControlNet.

\section{Task Definition}
\label{sec:task}

The focus of this work lies in improving the faithfulness of a generated image of height $H$ and width $W$ to a localized description.
The input for this task consists of 
(1) a prompt $X$ consisting of $N$ tokens,
(2) a collection of $R$ region masks $\{\mathbf{B}_r \}_{r = 0,\dots, R} $, where $\mathbf{B}_r \in \{0, 1\}^{H \times W}$\footnote{The region mask $\mathbf{B}_r$ contains the value 1 for pixels where the object should be present.}, and (3) region-token alignments $f_{\mathrm{RT}}: [1 \enleadertwodots N ] \rightarrow [ 0 \enleadertwodots R ]$ \footnote{Region 0 is the entire image, so a token assigned to region 0 is relevant everywhere in image.}
% $f_{\mathrm{RT}}: %N
% \af{X} \rightarrow R$ 
that specify which region each token in the text prompt belongs to. %where each mask $\mathbf{B}_r \in \{0, 1\}^{H \times W}$ is associated with a sub-sequence of tokens in $X$.
%First, a distinction must be made between \textit{global} and \textit{local} (or \textit{regional}) parts of the prompt: a global description describes the entire scene and subsets of its tokens are marked as corresponding to a certain region in the image.
As an example, consider the prompt ``\textit{\textcolor{black}{A photo of a \{\textcolor{blue}{blue crystal ball}\}$_1$, a \{\textcolor{red}{red tennis ball}\}$_2$ and a \{\textcolor{orange}{gold coin}\}$_3$ on a \{\textcolor{brown}{wooden table}\}$_4$}}'', where each colored sub-sequence is associated with the corresponding mask shown in \crefrange{fig:example:layers:1}{fig:example:layers:4}.

The goal is to generate an image where (1) object boundaries follow mask boundaries and (2) where the objects described by the %localized 
region-specific
parts of the prompts (i.e, the region descriptions) are generated in the parts of the output image associated with their region description.
The desired output for our example is given in \cref{fig:example:output}.

\section{Background}
%To introduce notation, 
In this section, we briefly review the diffusion-based text-to-image generation process, ControlNet, and some existing cross-attention control methods developed in the context of our task for diffusion models.

\subsection{Text-to-Image with Denoising Diffusion}
\label{sec:background:diffusion}

First, the textual description $X$ is encoded using the text encoder to produce the text embeddings 
% $c_{\mathrm{text}} = \{\vec{x}_n\}_{n=0..N}$, 
$\mathbf{X} = \{\vec{x}_n\}_{n=0..N}$, 
where each $\vec{x}_n\in \mathbb{R}^{d_x}$ and $\mathbf{X}$ can also be thought of as a tensor $\in \mathbb{R}^{N \times d_x}$. Here, $N$ is the number of input tokens and $d_x$ is the dimensionality of the embedding.
Stable Diffusion uses CLIP~\cite{clip} \afk{as encoder}, which is a transformer pre-trained on a text-image similarity task.

Then, a denoising model is used %$u$
to iteratively denoise an initial $z_T \sim \mathcal{N}(0, I)$ into an image $z_0$ using some solver, such as DDIM~\cite{ddim}.
At every iteration, the solver computes $z_{t-\delta} = s(u(z_{t}, t, \mathbf{X}), t, \delta)$, where $t$ is the denoising step, $u$ is the denoising model, $s$ is the solver algorithm, and $\delta$ is the step size.
The denoising model is typically implemented as a U-Net~\cite{unet},
%\af{[CHeck: Doenst $K$ need to enter in the function as well? And what is $c_{text}$? The same as $X$? I am not sure if the formular helps or rather confuses.]}
% that builds intermediate image feature maps $\mathbf{H}^{(l)}_{t}$ at every layer. %, \af{building on} downscaling and upscaling \af{operations} when changing resolutions. 
%In the case of text-to-image tasks, it 
\afk{which}
is conditioned on both the textual input $X$, %$c_{\mathrm{text}}$, 
%as well as
\afk{and}
the noisy image $z_t$.
%\af{[CHeck: Doenst $K$ need to enter in the function as well? And what is $c_{text}$? The same as $X$? I am not sure if the formular helps or rather confuses.]}

Conditioning on the text can be accomplished by using a cross-attention mechanism between 
the token embeddings $\mathbf{X} \in \mathbb{R}^{N \times d_x}$ and pixel-wise features $\mathbf{H} \in \mathbb{R}^{H \times W \times d_h}$:
%the $\vec{x}_i$'s and the (latent) pixel's embeddings $\vec{h}_{i,j}$. 
% We refer to the entire tensors of token embeddings and pixel-wise features as $\mathbf{X} \in \mathbb{R}^{N \times d_x}$ and $\mathbf{H} \in \mathbb{R}^{H \times W \times d_h}$, respectively.
% The (cross-) attention mechanism can then be briefly summarized as follows:
\begin{align}
\label{eq:attn}
    \mathbf{A} &= \mathrm{softmax} \Big( \dfrac{\mathbf{Q} \mathbf{K}^T }{\sqrt{d}} \Big) & ~~~ & 
    \mathbf{C} = \mathbf{A} \mathbf{V} \enspace .
\end{align}
Here, $\mathbf{Q} = f_Q (\mathbf{H}) \in \mathbb{R}^{H \times W \times d}$ are the query vectors computed by projecting the pixel-wise feature maps $\mathbf{H}$, %$\vec{h}_{i,j}$ \af{[Check: Are this vectors the entries of $\mathbf{H}$? And projekting on what?]}, 
and $\mathbf{K} = f_K(\mathbf{X}) \in \mathbb{R}^{N \times d}$ and $\mathbf{V} = f_V(\mathbf{X}) \in \mathbb{R}^{N \times d}$ are the key and value projections of the token embeddings, respectively.
$\mathbf{A} \in \mathbb{R}^{(H \cdot W) \times N}$ are the cross-attention scores.
Note that we omit layer and head indexes for clarity and that the dimensions $H$ and $W$ as well as the size of feature vectors $d$ and $d_h$ vary depending on the layer.
% \begin{align}
%     a_{i,j,n} &= \frac{\vec{q}_{i,j}\cdot \vec{k}_{n}}{\sqrt{d}} \\
%     \alpha_{i,j,n} &= \frac{e^{a_{i,j,n}}}{\sum_{n=0}^N e^{a_{i,j,n}}} \\
%     \vec{c}_{i,j} &= \sum_{n=0}^N \alpha_{i,j,n} \vec{v}_{n}      \enspace .
% \end{align}
% Here, $k_n$ and $v_n$ are the key and value projections of the textual embeddings $x_n$ and $q_{i,j}$ is the query projection of the pixel at position $(i,j)$ in the image input embedding $\vec{h} \in \mathbb{R}^{H \times W \times C}$.

\subsection{ControlNet}
\label{sec:background:controlnet}
ControlNet~\cite{controlnet} was recently proposed to improve control over the image composition. 
In addition to the prompt $X$,
%$c_{\mathrm{text}}$ \af{[Check: Is this the same as the textual desciption $X$ before?]}, 
ControlNet expects an image $c_{\mathrm{img}}$ as part of the input for the generation process.
In order to incorporate conditioning based on $c_{\mathrm{img}}$, first a control model is defined that copies the down-sampling and middle blocks of the latent diffusion model's U-Net. 
The control model also contains an additional block of convolutional layers that encodes the control signal $c_{\mathrm{img}}$ and is trained from scratch.
% The latent features of the noisy image $x_t$ are merged with features extracted from $c_{\mathrm{img}}$ before feeding them into the copied down-sampling block.
% The features computed by the control model are then added to the features computed by its sibling in the main U-Net before feeding them into the up-sampling blocks of the main U-Net.
% The authors introduce zero-convolutions to control how much information is being added to the original features.
% The features computed by the control model are then added to the features computed by its sibling in the main U-Net before feeding them into the up-sampling blocks of the main U-Net.
The features computed by the control model are added to the features computed by its sibling in the main U-Net before feeding them into the up-sampling blocks of the main U-Net. 
We refer the reader to Supplement~\ref{sec:supp:controlnet} and to the original work of 
~\citet{controlnet} for
a more detailed explanation.
ControlNet supports different types of conditioning input, such as segmentation maps, depth maps or human pose.
Each %of these control methods 
\afk{type}
requires the training of a separate control model dedicated to that type of conditioning. 
Combining several control signals~\cite{unicontrolnet} is an active research area.

Note that while ControlNet allows us to control the image layout using segmentation maps, it lacks a mechanism to precisely control what object is generated inside each region.
\afk{As a consequence,} when faced with ambiguous layouts or improbable region assignments, plain ControlNet can not correctly process the prompt, \afk{as illustrated in our qualitative study in \cref{sec:qualitative}.}

\subsection{Cross-attention control}
\label{sec:cac}
Modifying cross-attention~\cite{attention} scores in the transformer~\cite{transformer} blocks of the U-Net can provide a degree of spatial control and attribute assignment.
Here we %formally describe the general mechanism before discussing nuances of the 
give a brief introduction to the
different previously proposed methods \afk{for cross-attention control} that we investigate \afk{in this work}.

In addition to the token embeddings $\vec{x}_n$, also the following inputs are expected: (1) the region masks $\mathbf{B}_r \in \{0, 1\}^{H \times W}, r \in [ 1 \enleadertwodots R ]$ for each region % $r$ in the set of provided regions $R$ 
and (2) region-token alignments $f_{\mathrm{RT}}: [ 1 \enleadertwodots N ] \rightarrow [ 0 \enleadertwodots R ]$.
In general, the cross-attention control mechanism simply projects the region-token alignments onto the cross-attention scores and manipulates them to stimulate cross-attention from the specified region to the corresponding set of tokens and to prevent from attending to other localized descriptions.

\newcommand{\W}{\mathbf{W}}

%\subsubsection{eDiff-I}
\paragraph{eDiff-I (community edition)}
The first cross-attention control method we consider is a re-implementation~\cite{ediffipp} %based on 
%\af{from~\citet{ediffipp}} 
of the approach proposed \afk{by \citet{ediffi}}.
%We reimplement it for Stable Diffusion with ControlNet and further refer to it as \textbf{eDiff-I}.
It takes the region-annotated prompt and the region masks, and forces cross-attention to attend to certain words from the corresponding regions by modifying the cross-attention scores\footnote{Note that this formulation by the community slightly differs from the one proposed by~\citet{ediffi}. We use this formulation since in our early experiments, we found it to perform slightly better.}:
% $W$ to the cross-attention score, which is defined by
\begin{align}
    \mathbf{A} &= \mathrm{softmax} \Big( \W + \dfrac{\mathbf{Q} \mathbf{K}^T }{\sqrt{d}} \Big) \enspace \label{eq:attnmod} \\
  \W &= W' \cdot \log (1 + \sigma^2) \cdot \mathrm{std} (\mathbf{Q} \mathbf{K}^T) \cdot \mathbf{B}_{f_{\mathrm{RT}}(n)}  \enspace .
    % W &= W' \cdot \log (1 + \sigma) \cdot \max (\mathbf{Q} \mathbf{K}^T)   \enspace .   
\end{align}
%here 
Here, $\W$ is scheduled to decrease as the denoising process progresses using $\sigma$, which is a scalar specifying the current noise level.
% computed as $\sigma = \sqrt{\frac{1 - \hat{\alpha}}{\hat{\alpha}}}$. 
$W'$ is a hyper-parameter controlling the overall degree of attention change.
% The modified cross-attention equation are then given by 
% \begin{align}
%     \mathbf{C} &= \mathrm{softmax} \Big( W \cdot \mathbf{B}_{f_{\mathrm{RT}}(n)}  + \dfrac{\mathbf{Q} \mathbf{K}^T }{\sqrt{d}} \Big) \mathbf{V} \enspace . %\\
%  %   W &= W' \cdot \log (1 + \sigma^2) \cdot \mathrm{std} (\mathbf{Q} \mathbf{K}^T)   \enspace 
%     % W &= W' \cdot \log (1 + \sigma) \cdot \max (\mathbf{Q} \mathbf{K}^T)   \enspace ,
% \end{align}
%Note that this formulation by the community slightly differs from the one proposed \afk{by~\citet{ediffi}.}
%We use this formulation since in our early experiments, we found it to perform slightly better.

% We also found a community implementation~\cite{ediffipp}
% %that we further refer to as \textbf{eDiff-I++} 
% that changes the weight schedule slightly and use this implementation for our analysis since in our early experiments, we found it to perform slightly better.
% \begin{align}
%     W &= W' \cdot \log (1 + \sigma^2) \cdot \mathrm{std} (\mathbf{Q} \mathbf{K}^T)   \enspace 
% \end{align}

\paragraph{CAC}
%Rather than

Instead of boosting cross-attention scores between regions and their descriptions in the prompt, \citet{cac} propose to apply a binary mask that eliminates attention between regions and non-matching region descriptions.
The binary mask is applied \emph{after} softmax normalization\footnote{Since the source code for~\cite{cac} has not been made available at the time of this writing, we had to rely on the descriptions given in the paper for our implementation.}:
% \afk{Instead of}
% modifying the (unnormalized) cross-attention scores, \citet{cac} % instead 
% propose 
% to apply the binary mask :
\begin{align}
    \mathbf{A} &= \mathrm{softmax} \Big( \dfrac{\mathbf{Q} \mathbf{K}^T }{\sqrt{d}} \Big)  \odot \mathbf{B}_{f_{\mathrm{RT}}(n)}  \enspace.
\end{align}
Note that the attention weights are no longer normalized after applying the mask.

\paragraph{DenseDiffusion}
\afk{\citet{densediffusion}} present another variant of cross-attention control where attention scores for the tokens describing a region are increased while attention scores to other tokens are decreased.
In addition, the method also proposes to scale the degree of change by the region size fraction $S$ and uses a schedule that decreases polynomially. 
Cross-attention scores are modified by redefining $\W$ from Eq.~\ref{eq:attnmod} as follows:
%
% \begin{align}
%     \mathbf{C} =&~ \mathrm{softmax} \Big(    \dfrac{   \mathbf{Q} \mathbf{K}^T    + W  }{  \sqrt{d}  }    \Big) \mathbf{V} \enspace \,\,\, \text{with} \\
%     W =& ~ W' \cdot (1 - S) \cdot \Big(   \dfrac{(T-t)}{T}   \Big) ^5 \label{eq:dd2}  \nonumber
%     \\ & \cdot ( \mathbf{B}_{f_{\mathrm{RT}}(n)} \cdot \mathbf{M}_{\mathrm{+}}
%    - (1-\mathbf{B}_{f_{\mathrm{RT}}(n)}) \cdot  \mathbf{M}_{\mathrm{-}} )  \enspace ,  
% \end{align}
\begin{align}
    \W =& ~ W' \cdot \big(  \dfrac{t}{T}  \big) ^5 \cdot (1 - S) \nonumber \\ &
    \cdot ( \mathbf{B}_{f_{\mathrm{RT}}(n)} \odot \mathbf{M}_{\mathrm{+}}
   - (1-\mathbf{B}_{f_{\mathrm{RT}}(n)}) \odot  \mathbf{M}_{\mathrm{-}} )
\end{align}
%
%\begin{align}
%  W = &~ W' \cdot (1 - S) \cdot \Big(   \dfrac{(T-t)}{T}   \Big) ^5  \nonumber
%    \\ & \cdot ( \mathbf{B}_{f_{\mathrm{RT}}(n)} \cdot \mathbf{M}_{\mathrm{+}}
%    - (1-\mathbf{B}_{f_{\mathrm{RT}}(n)}) \cdot  \mathbf{M}_{\mathrm{-}} )  \enspace ,    
%\end{align}
where $\mathbf{M}_{\mathrm{+}}$ and $\mathbf{M}_{\mathrm{-}}$ specify the maximum increase and decrease for every token:
\begin{align}
    \mathbf{M}_{\mathrm{+}} =& ~ \max (\mathbf{Q} \mathbf{K}^T) - \mathbf{Q} \mathbf{K}^T \enspace , \\
    \mathbf{M}_{\mathrm{-}} =& ~ \mathbf{Q} \mathbf{K}^T - \min (\mathbf{Q} \mathbf{K}^T) \enspace .
\end{align}
In addition to cross-attention control, DenseDiffusion~\cite{densediffusion} also includes self-attention control using a similar method.
%However, we do not experiment with self-attention control since we found it to under-perform ControlNet in our early experiments.

\section{Approach}
\label{sec:method}

In this work, we use the segmentation-based ControlNet on top of Stable Diffusion as it already provides us with a means to control image layout with high precision by specifying segmentation maps.
Note that even though the image layout is controlled, it still leaves the model with freedom how to assign the objects mentioned in the text prompt to the regions.
%The core of our work is %investigating
%afk{the investigation of}
The core of our work is to investigate cheap methods to enable ControlNet to adhere to localized descriptions in order to solve the task described in Section \ref{sec:task}. 
To do so, we (1) adapt existing efficient training-free cross-attention control methods for ControlNet, and (2) propose a novel cross-attention control technique that facilitates
%efficient training-free cross-attention control methods to facilitate 
better object and attribute alignment between the text prompt and the generated image, while minimizing image artifacts due to the introduction of cross-attention control.
% \dl{and (3) investigate fine-tuning strategies for ControlNet to foster overall performance [TODO: im not sure i'd add this because we don't evaluate it]}. 

\subsection{Cross-Attention Control in ControlNet}

\afk{In a first attempt to enable ControlNet to solve task defined in Section \ref{sec:task}, we adapt and integrate several representative}
%Firstly, we adapt and compare several representative 
cross-attention control methods into ControlNet. % in an attempt to improve its fidelity to the localized prompt and provided regions.
\afk{More precisely,} we implement the methods mentioned in \cref{sec:cac}, and apply cross-attention control in both the control network as well as the main diffusion U-Net.
%Through the analysis presented in this work, we identify shortcomings of these methods that prevent them from satisfying more challenging generation cases.
%
Note that different layers of the U-Net work at different resolutions as the network consists of a stack of down-scaling layers, followed by up-scaling layers.
%\af{[Check: Is the following sentence a desciption of what we do, or how it is generally done? In the first case, we should shift this into our approach.]}
We simply up- and down-scale the mask $\mathbf{B}_r$ as necessary using bilinear interpolation.

\subsection{Attention redistribution}
\label{sec:transferandboost}
The previously discussed cross-attention control methods have certain shortcomings.
Firstly, most methods are very sensitive to the selection of the time steps during which attention manipulation is performed and to what degree \afk{it is performed}.
Most methods studied here rely on the assumption that the image layout is determined
in the initial denoising steps and do not modify attention in later generation stages. 
Such methods (e.g., DenseDiffusion and eDiff-I) therefore place a high degree of control in the initial stages of decoding and quickly drop it to near-zero values by roughly $t=750$ (%note that 
\afk{if}
generation starts with $t=T=1000$).
However, this procedure can still lead to concept bleeding in highly ambiguous cases, for example when generating objects of similar shapes and color.
% However, this \afk{procedure has} %can also have 
% disadvantages in cases \afk{where control throughout the whole generation process is beneficial}, for example when generating objects of similar shapes and color.
After the initial heavily controlled stage, the model becomes uncontrolled and fine details such as object texture can no longer be clearly assigned when multiple similar objects are present.
We confirm this in our qualitative experiments in Sec.~\ref{sec:qualitative}.
This shows that is desirable to have an attention control method that remains active throughout the denoising process while still minimizing image quality degradation.

A second consideration is that at different heads of different layers and at different generation stages, the attention weights behave differently and indiscriminate boosting of attention can lead to a decrease in image quality and a higher sensitivity to the attention control schedule.
The exception is CAC, which disables attention to the descriptions of irrelevant regions throughout the entire generation process.

Thirdly, when simply disabling attention to irrelevant tokens, like CAC, the attention ``mass'' is either mostly transfered to the most probable tokens or is lost. 
This can be problematic when the initial random image $x_T$ leads the model to mostly attend to the wrong region descriptions, in this case, attention to the wrong regions is dropped but the attention weights to the correct region remain at their initial (possibly low) values.
In addition, in CAC-style control, the attention weights no longer sum to one.
%, which could be another source of image degradation.
In our experiments, we observe that for CAC-style control, small objects are sometimes not generated and the grounding behavior is more inconsistent across seeds, where some region descriptions are ignored, as illustrated in the qualitative study in \cref{sec:qualitative}.

\begin{figure}[t]
    \centering
    \includegraphics[width=0.99\linewidth]{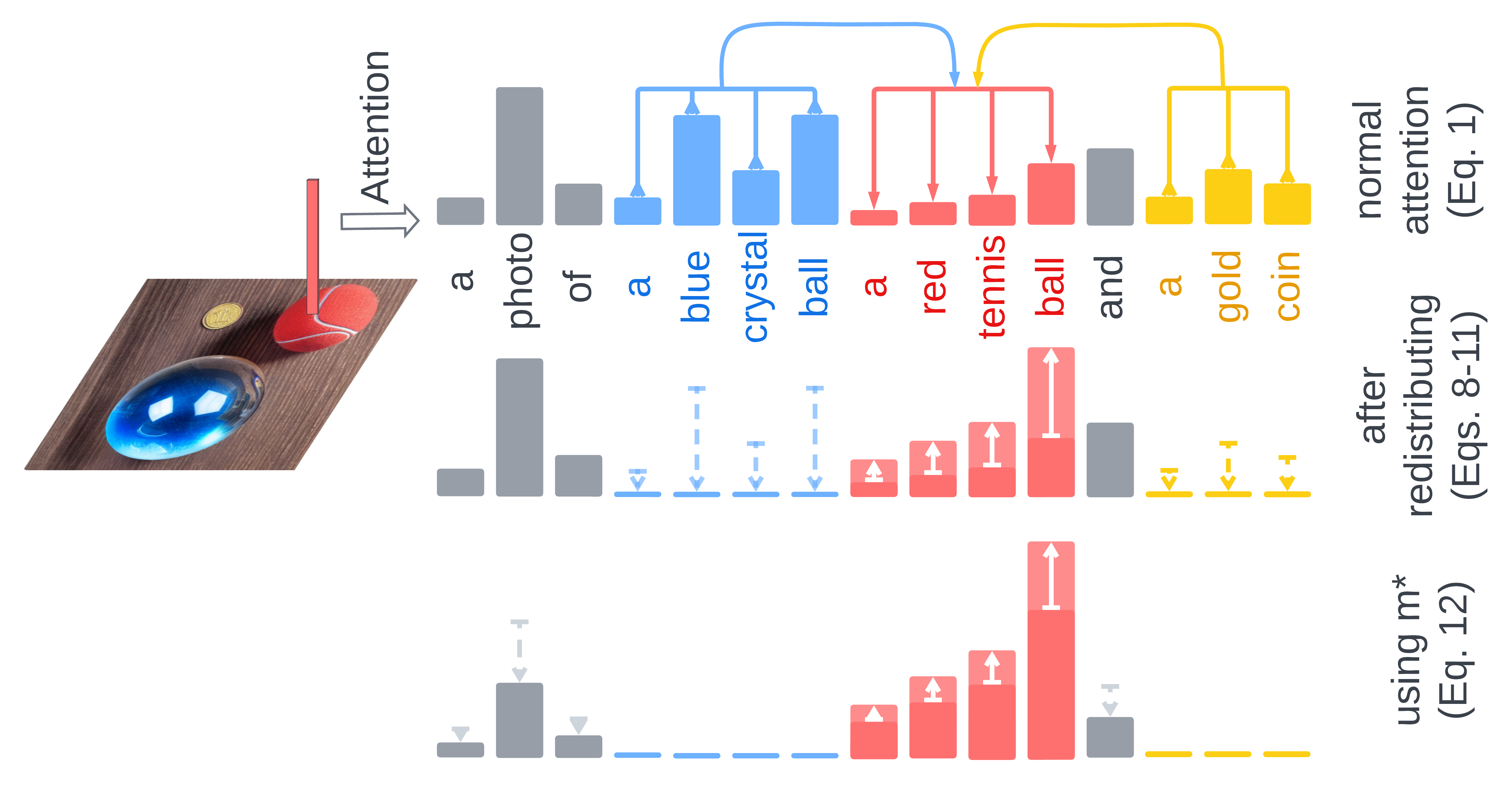}
    \caption{A diagram illustrating attention redistribution and attention boosting with the running example.
    }
    \label{fig:diagram}
\end{figure}

\paragraph{CA-Redist:} To address these shortcomings, we propose a cross-attention manipulation method that we refer to as \textbf{cross-attention redistribution (CA-Redist)} and that redistributes attention from irrelevant region descriptions to the relevant one.
Concretely, this is accomplished by (1) computing the total amount of region-specific attention $m$, which can vary across heads and layers, (2) separately normalizing region-specific and region-agnostic attention weights to obtain $\mathbf{A}_{\mathrm{local}}$ and $\mathbf{A}_{\mathrm{global}}$, respectively, and (3) mixing the two resulting attention distributions using $m$.
Note that this is done separately for every pixel,
%Here, it 
\afk{and that it}
is assumed that every pixel in the image is assigned to exactly one region description. % in the prompt.
Fig.~\ref{fig:diagram} provides an illustration of CA-Redist.
This method is defined as follows:
\begin{align}
    % \mathbf{A} &= \mathrm{softmax} \Big( \dfrac{\mathbf{Q} \mathbf{K}^T }{\sqrt{d}} \Big) \mathbf{V}  \enspace    , 
    \mathbf{A} &= m \cdot \mathbf{A}_{\mathrm{local}} + (1 - m) \cdot \mathbf{A}_{\mathrm{global}}   \enspace    , \label{eq:caredist1} \\
    \mathbf{A}_{\mathrm{local}}    &=      \mathrm{softmax} \Big(     \log (   \mathbf{B}_{f_{\mathrm{RT}}(n)}   )    +     \dfrac{\mathbf{Q} \mathbf{K}^T    }{\sqrt{d}} \Big)  \enspace    ,    \label{eq:caredist2} 
\end{align}
\begin{align}
    \mathbf{A}_{\mathrm{global}}    &=      \mathrm{softmax} \Big(     \log (   1 - \mathbf{B}_R   )    +     \dfrac{\mathbf{Q} \mathbf{K}^T    }{\sqrt{d}} \Big) \enspace    ,    \label{eq:caredist3} \\
    m &= \sum_{n=0}^N \mathbf{A}_{n} \cdot  \mathbf{B}_{R,n}    \enspace   ,\label{eq:caredist4}
\end{align}
where $\mathbf{A}_n$ is the attention weight corresponding to the $n$-th token, as computed by Eq.~\ref{eq:attn} and
$\mathbf{B}_R$ is a mask that is set to one for all tokens that belong to \emph{any} region description.
Thus, $\mathbf{A}_{\mathrm{global}}$ computes an attention distribution over all tokens except those in any region description and $\mathbf{A}_{\mathrm{local}}$ is zero everywhere except the correct region description.
The mixture between the two makes sure to retain the same attention weights for the non-region tokens (in other words, keeping $\mathbf{A}_{\mathrm{global}} \approx \mathbf{A}$ for tokens where $\mathbf{B}_R$ is zero).

The attention to relevant region-specific parts of the prompt can further be increased by replacing $m$ with $m^*$ as defined below, where $m$ can be modified in two ways, using hyper-parameters $W_m \geq 0$ and $W_a \geq 0$ that boost the attention to relevant parts of the prompt multiplicatively or additively, respectively.
\begin{align}
    m^* =&~ \min \Big(1, \max \big( 0, \\  &   m \cdot (1 + W_m \cdot W'') + W_a \cdot W'' \cdot (1 - S)   ~~~~ \big)\Big)        \enspace    , \nonumber 
\end{align}
%\af{[Check: This I do not concretely understand. Should $m*$ replace $m$ in equation 11?. And why does this correspond to "two ways of changig m"?}
where $S$ is the fraction of the surface area that a region occupies in the image (same as defined for DenseDiffusion) and $W''$ specifies the schedule of attention boost in CA-Redist and depends on the current denoising step $t$:
\begin{align}
    W'' &= \begin{cases}
        1 & \text{if $t \geq T_{\mathrm{s}}$} \\
        \frac{1}{2} + \frac{1}{2} \sin (\pi \cdot \frac{t - T_{\mathrm{thr}}}{T_{\mathrm{s}} -  T_{\mathrm{e}}})   & \text{if $T_{\mathrm{s}} > t > T_{\mathrm{e}}$} \\
        0 & \text{if $T_{\mathrm{e}}$} \geq t  
    \end{cases} \label{eq:cosineschedule} \\
    T_{\mathrm{s}} &= T_{\mathrm{thr}} + RT/2 \quad \quad \quad
    T_{\mathrm{e}} = T_{\mathrm{thr}} - RT/2
\end{align}
This schedule is controlled by the threshold step $T_{\mathrm{thr}} \in [1 \enleadertwodots T ]$ and the threshold softness $R \in [0, 1]$.
Unless otherwise specified, in our experiments, we set $T_{\mathrm{thr}} = T$ (see also Supplement~\ref{sec:supp:eq15}).

% \begin{align}
%     W'' &= \begin{cases}
%         1 & \text{if $t \geq T_{\mathrm{thr}} + \frac{RT}{2}$} \\
%         0 & \text{if $t \leq T_{\mathrm{thr}} - \frac{RT}{2}$}  \\
%         \frac{1}{2} + \frac{1}{2} \sin (\pi \cdot \frac{t - T_{\mathrm{thr}}}{T_{\mathrm{s}} -  T_{\mathrm{e}}})   & \text{otherwise} \\
%     \end{cases}  \label{eq:cosineschedule}
% \end{align}
% \begin{align}
%     W'' &= \begin{cases}
%         1 & \text{if $t \geq T_{\mathrm{s}}$} \\
%         \frac{1}{2} + \frac{1}{2} \sin (\pi * \frac{t - T_{\mathrm{thr}}}{T_{\mathrm{s}} -  T_{\mathrm{e}}})   & \text{if $T_{\mathrm{s}} > t > T_{\mathrm{e}}$} \\
%         0 & \text{if $T_{\mathrm{e}}$} \geq t  \\
%     \end{cases} \\
%     T_{\mathrm{s}} &= T_{\mathrm{thr}} + T * R/2\\
%     T_{\mathrm{e}} &= T_{\mathrm{thr}} - T * R/2
% \end{align}
% This schedule is controlled by the threshold step $T_{\mathrm{thr}} \in [1 \enleadertwodots T ]$ and the threshold softness $R \in [0, 1]$.
% Note that $t$ starts from $T$ so attention boost is active more in the initial stages of denoising.
%
Multiplicative manipulation using $W_m$ is stronger for heads with higher attention weights to region descriptions and remains low for those that attended to tokens outside of any region description.
Additive manipulation using $W_a$ forces attention to increase to region-specific tokens for all heads, similarly to previous approaches.

%We refer to this method as \textbf{\afk{cross-attention redistribution} (CA-Redist)}.

\subsection{ControlNet fine-tuning}
\label{sec:controlnetmod}
%Even though \af{the} segmentation map-based ControlNet appears to function well, 
\afk{We decided to further fine-tune the segmentation map-based ControlNet, since it} %due to the following consideration
%\af{The} segmentation map-based ControlNet 
was pre-trained using COCO~\cite{coco} and ADE20K~\cite{ade40k,ade40k2} semantic segmentation maps, where certain colors are associated with certain object classes.
In our application scenario, however, we are not interested in using the semantic segmentation conditioning, but rather conditioning image inputs similar to panoptic segmentation, in order to clearly separate objects of the same class and support a randomized color map since we don't have access to the class of every object we want to generate. 

\paragraph{Fine-tuning with custom conditioning colors:}
To transition from semantic segmentation to panoptic-like segmentation, we fine-tuned parts of the segmentation-based ControlNet v1.1 model on COCO2017 panoptic segmentation data, while replacing colors in the original annotation images with randomized colors.
Only the part of the ControlNet model was fine-tuned that extracts features from the conditioning image before adding them to the first-layer features of the latent image $z_t$.
We found this to slightly increase image quality but did not further affect model behavior.

\paragraph{Text-controlled ControlNet baseline:}
We also implemented and trained a ControlNet variant where in addition to the segmentation input, the model also receives pooled CLIP encodings of short region descriptions (e.g. ``\textit{dog}'') as part of the conditioning input, in the hopes that the model learns to use the additional text features in order to select the relevant parts of the original prompt (which contains the full region description, e.g. ``\textit{brown german shepherd}'').
However, in our experiments with training on COCO2017, this did not result in a model that respects region descriptions better.
For this reason, this baseline is not included in our experiments and further investigation is left for future work.

\begin{figure}[t]
    \centering
    \includegraphics[width=0.85\linewidth]{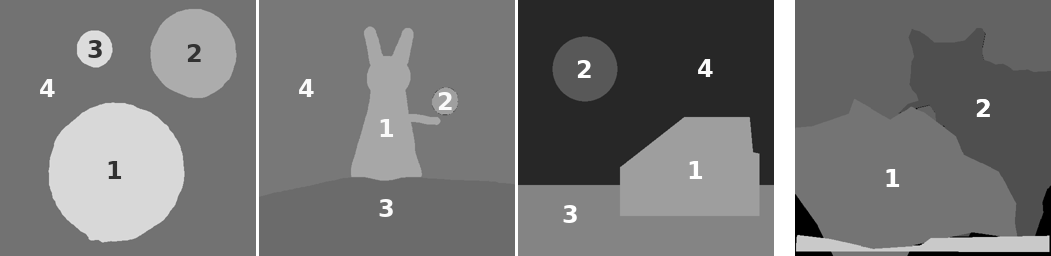}
    \caption{Layouts used for qualitative comparison throughout this paper (first three layouts are used in Fig.~\ref{fig:qualitative}, the last layout in Fig.~\ref{fig:complexshapes}).
    }
    \label{fig:layouts}
\end{figure}

\begin{figure*}[!htb]
    \centering
    \includegraphics[width=0.95\textwidth]{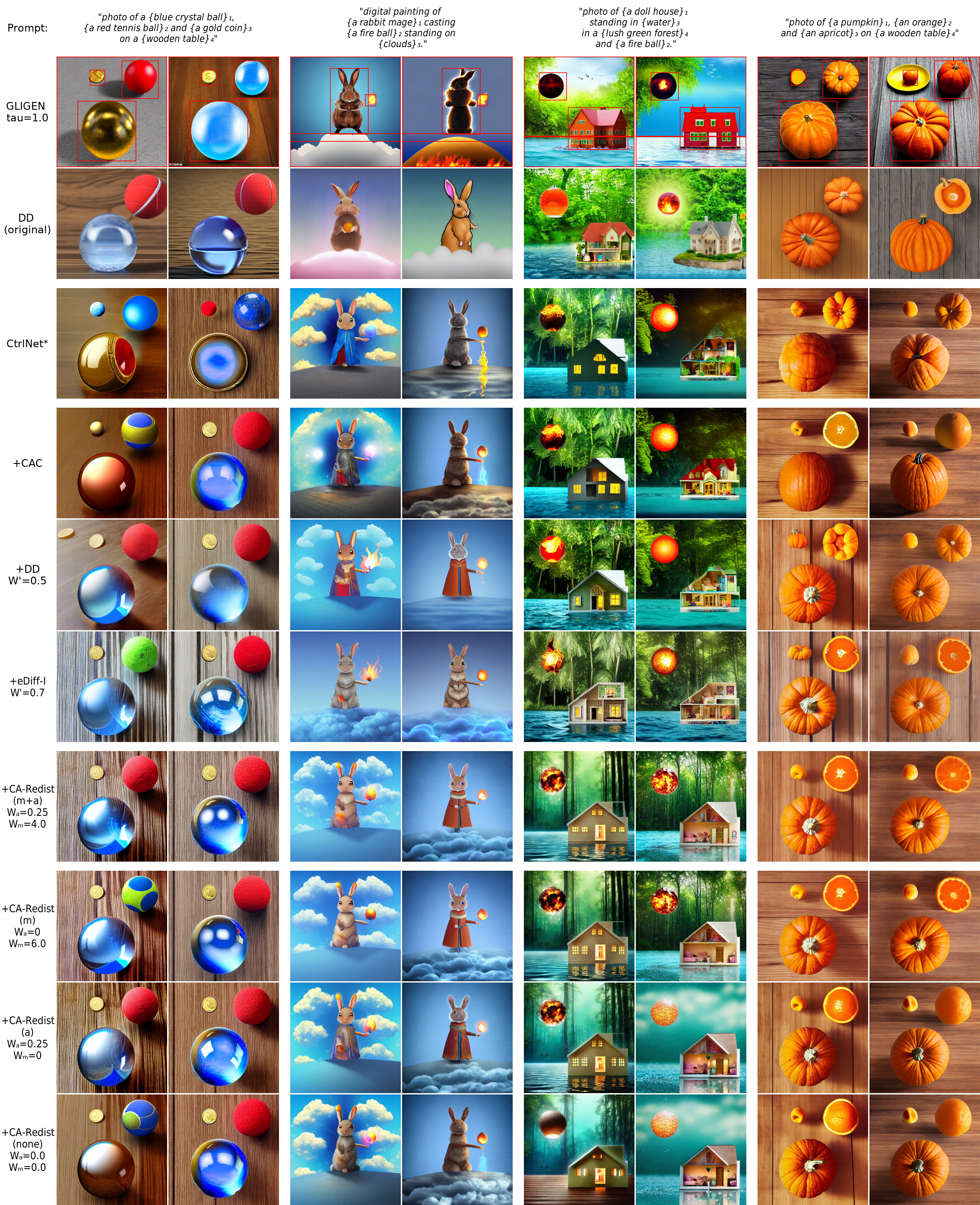}
    \caption{A qualitative comparison of different cross-attention control methods in ControlNet-extended Stable Diffusion. 
    % All image in one column are generated using the same initial seed and conditioning image across all experiments.
    Results for multiple seeds are shown to illustrate how consistent the generation results are.
    See Fig.~\ref{fig:layouts} for layout specification.
    }
    \label{fig:qualitative}
\end{figure*}

\section{Experiments}
\label{sec:experiments}

We compare the following methods of cross-attention control on top of the (lightly fine-tuned) ControlNet:
%(1) only ControlNet: a (lightly fine-tuned) baseline without any attention control, 
(1) \afk{eDiff-I}, %: % the method outlined in~\cite{ediffi}, in its improved implementation by the community, 
(2) CAC,
(3) DenseDiffusion (DD),
and (4) CA-Redist.
We also compare against %plain ControlNet as well as 
the original implementations of GLIGEN~\footnote{As provided in the Huggingface Diffusers library} and DenseDiffusion~\footnote{\url{https://github.com/naver-ai/densediffusion}} as points of reference of related work that does not rely on ControlNet.
Note that we used the main variant of GLIGEN that takes as input bounding boxes and localized descriptions.\footnote{GLIGEN also provides a variant that takes a segmentation map as conditioning but it does not use localized textual descriptions so it is unfit for our task.}
Comparison with SceneComposer~\cite{scenecomposer} and SpaText~\cite{spatext} was not possible because the code and data have not been publicly released at the time of this writing. 
It must also be noted that both these approaches have been extensively trained, whereas our proposed method also works training-free with segmentation- and scribble-based ControlNet. 
In our experiments, we did minimal fine-tuning of a small part of ControlNet on a readily available dataset to align color schemes since we found it slightly improved image quality in our early experiments.

Details on the experimental setup can be found in \cref{sec:supp:expsetup} in supplementary material.

\subsection{Qualitative study}
\label{sec:qualitative}

A qualitative comparison of the different attention control methods is presented in \cref{fig:qualitative}.
The layouts used are specified in \cref{fig:layouts}, where the numbers correspond to the numbered phrases in the prompts in \cref{fig:qualitative}.
For a qualitative comparison using ControlNet trained for sketch conditioning (scribbles), please consider Supplement~\ref{sec:supp:scribbles}.

\paragraph{Comparison of baselines:}
\label{sec:qualitative:baselines}
The baselines that don't rely on ControlNet appear to fail at the task with challenging inputs.
In fact, \textbf{GLIGEN}~\cite{gligen}~\footnote{Note that we used bounding boxes for conditioning, so exact layout is not expected to match.}, despite enabling its adapter throughout the entire generation process ($\tau = 1.$), mostly failed to assign the right textures and colors to objects for the challenging prompts involving multiple round objects.
This can be speculated to be attributable to the fact that during training, GLIGEN is insufficiently exposed to such challenging examples and doesn't learn to take into account localized descriptions in later generation steps.

In comparison, the original implementation of \textbf{DenseDiffusion}~\cite{densediffusion} is better at assigning objects to regions.
However, we see it fail in examples where the the shape and colors in the early generation stages are ambiguous, as illustrated in the last two columns of Fig.~\ref{fig:qualitative}.
Also, it seems to ignore smaller object masks and does not adhere to mask boundaries as precisely as ControlNet (however, it must be noted that DenseDiffusion is completely training-free).

Finally, \textbf{plain ControlNet}, pre-trained with semantic segmentation labels, and fine-tuned on COCO2017 data for panoptic segmentation with randomized colors, can already assign the correct description to the correct region if the mask shapes are distinctive enough.
This is illustrated in the fifth column, where all shapes can be unambiguously matched with a region description (e.g. ``fire ball'' is a circular shape,  ``doll house'' is a trapezoid shape).
However, when faced with ambiguity in layout specification (e.g. three circles), plain ControlNet randomly assigns objects and colors and suffers from concept bleeding (for example, assigning ``\textit{gold}'' to a ball whereas we described a gold coin).
Additionally, it can struggle when faced with improbable descriptions, such as a rabbit mage standing on clouds. %, where clouds take on a not cloudlike texture and/or color.

\begin{table*}[]
    \centering
    \small
    \begin{tabular}{l  c c c  c c}
    \toprule
    & BRISQUE $\downarrow$ & MANIQA $\uparrow$ & LAION Aest $\uparrow$ & LocalCLIP Logits $\uparrow$ & LocalCLIP Prob. $\uparrow$ \\
    \midrule
%%% INSERT HERE
GLIGEN ($\tau=1.0$) & $30.56$ $^{\tiny \pm 1.78}${\scriptsize ($10.23$) } & $0.65$ $^{\tiny \pm 0.00}${\scriptsize ($0.70$) } & $5.51$ $^{\tiny \pm 0.04}${\scriptsize ($6.08$) } & $21.60$ $^{\tiny \pm 0.17}${\scriptsize ($23.19$) } & $0.33$ $^{\tiny \pm 0.01}${\scriptsize ($0.50$) } \\
DD (original) & $32.69$ $^{\tiny \pm 1.14}${\scriptsize ($12.96$) } & $0.65$ $^{\tiny \pm 0.00}${\scriptsize ($0.69$) } & \cellcolor{lightgray!75}$5.76$ $^{\tiny \pm 0.03}${\scriptsize ($6.33$) } & $21.52$ $^{\tiny \pm 0.06}${\scriptsize ($22.83$) } & $0.45$ $^{\tiny \pm 0.01}${\scriptsize ($0.58$) } \\
ControlNet* & $25.23$ $^{\tiny \pm 1.46}${\scriptsize ($7.65$) } & $0.66$ $^{\tiny \pm 0.00}${\scriptsize ($0.70$) } & $5.70$ $^{\tiny \pm 0.02}${\scriptsize ($6.25$) } & $20.99$ $^{\tiny \pm 0.12}${\scriptsize ($22.60$) } & $0.25$ $^{\tiny \pm 0.01}${\scriptsize ($0.41$) } \\
\quad +CAC & $26.97$ $^{\tiny \pm 1.02}${\scriptsize ($8.99$) } & $0.67$ $^{\tiny \pm 0.00}${\scriptsize ($0.71$) } & $5.71$ $^{\tiny \pm 0.05}${\scriptsize ($6.29$) } & $22.28$ $^{\tiny \pm 0.23}${\scriptsize ($23.84$) } & $0.44$ $^{\tiny \pm 0.02}${\scriptsize ($0.61$) } \\
\quad +DD (w=0.5) & \cellcolor{lightgray!35}$24.50$ $^{\tiny \pm 1.94}${\scriptsize ($8.39$) } & $0.66$ $^{\tiny \pm 0.00}${\scriptsize ($0.70$) } & \cellcolor{lightgray!35}$5.74$ $^{\tiny \pm 0.05}${\scriptsize ($6.21$) } & $22.93$ $^{\tiny \pm 0.11}${\scriptsize ($24.18$) } & $0.48$ $^{\tiny \pm 0.02}${\scriptsize ($0.59$) } \\
\quad +eDiff-I (w=0.5) & \cellcolor{lightgray!75}$23.75$ $^{\tiny \pm 1.15}${\scriptsize ($10.86$) } & $0.66$ $^{\tiny \pm 0.00}${\scriptsize ($0.70$) } & $5.73$ $^{\tiny \pm 0.02}${\scriptsize ($6.22$) } & $23.36$ $^{\tiny \pm 0.13}${\scriptsize ($24.43$) } & \cellcolor{lightgray!35} $0.58$ $^{\tiny \pm 0.02}${\scriptsize ($0.68$) } \\
\quad +CA-Redist (m+a) & $25.40$ $^{\tiny \pm 2.29}${\scriptsize ($10.10$) } & $0.67$ $^{\tiny \pm 0.00}${\scriptsize ($0.71$) } & \cellcolor{lightgray!35}$5.74$ $^{\tiny \pm 0.02}${\scriptsize ($6.22$) } & \cellcolor{lightgray!75} $23.77$ $^{\tiny \pm 0.11}${\scriptsize ($24.89$) } & \cellcolor{lightgray!75} $0.62$ $^{\tiny \pm 0.01}${\scriptsize ($0.73$) } \\
\quad +CA-Redist (m) & $27.37$ $^{\tiny \pm 2.00}${\scriptsize ($10.54$) } & $0.68$ $^{\tiny \pm 0.00}${\scriptsize ($0.71$) } & $5.68$ $^{\tiny \pm 0.05}${\scriptsize ($6.18$) } & \cellcolor{lightgray!35} $23.70$ $^{\tiny \pm 0.13}${\scriptsize ($24.80$) } & \cellcolor{lightgray!75} $0.62$ $^{\tiny \pm 0.01}${\scriptsize ($0.72$) } \\
\quad +CA-Redist (a) & \cellcolor{lightgray!35}$24.86$ $^{\tiny \pm 1.33}${\scriptsize ($8.41$) } & $0.67$ $^{\tiny \pm 0.00}${\scriptsize ($0.71$) } & $5.69$ $^{\tiny \pm 0.01}${\scriptsize ($6.20$) } & $23.52$ $^{\tiny \pm 0.07}${\scriptsize ($24.74$) } & \cellcolor{lightgray!35} $0.58$ $^{\tiny \pm 0.01}${\scriptsize ($0.70$) } \\
\quad +CA-Redist (none) & $25.13$ $^{\tiny \pm 1.54}${\scriptsize ($8.14$) } & $0.67$ $^{\tiny \pm 0.00}${\scriptsize ($0.70$) } & $5.65$ $^{\tiny \pm 0.04}${\scriptsize ($6.20$) } & $23.26$ $^{\tiny \pm 0.04}${\scriptsize ($24.60$) } & $0.56$ $^{\tiny \pm 0.01}${\scriptsize ($0.70$) } \\

%%% END INSERT

\bottomrule
    \end{tabular}
    \caption{Comparison with image quality and localized prompt adherence metrics using our \textsc{SimpleScenes} dataset. The arrows  indicate whether higher ($\uparrow$) or lower ($\downarrow$) is better. The numbers are reported in the format $\mathrm{MEAN}^{\pm \mathrm{STD}} (\mathrm{BEST})$, obtained using five different seeds.
    }
    \label{tab:quanteval}
\end{table*}

\begin{table}[]
    \centering
    \small
    \begin{tabular}{l  c c c c c}
    \toprule
    & FID $\downarrow$ & KID ($\times 10^3)$ $\downarrow$ \\% & CLIP-FID $\downarrow$ \\ %&  Clean FID & Clean KID \\
    \midrule
%%% INSERT HERE
GLIGEN ($\tau=1.0$) & \cellcolor{lightgray!75} 23.84 &  \cellcolor{lightgray!75} 4.289 \\% & 6.49 \\ %& 23.91 & 4.243 \\
DD (original)        & 37.68 &  7.923 \\% & 18.73 \\ %& 37.94 & 7.923 \\
ControlNet* & 28.84 &  5.150 \\% & 16.02 \\ %& 29.14 & 5.229 \\
\quad +CAC & 27.42 &  4.916 \\% & 14.99 \\ %& 27.72 & 5.037 \\
\quad +DD (w=0.5) & 28.30 &  5.422 \\% & 16.78 \\ %& 28.52 & 5.521 \\
\quad +eDiff-I (w=0.5) & 28.72 &  6.074 \\% & 17.77 \\ %& 28.91 & 6.076 \\
\quad +CA-Redist (m+a) & 27.11 &  5.276 \\% & 17.10 \\ %& 27.35 & 5.429 \\
\quad +CA-Redist (m) & 27.78 &  5.547 \\% & 17.33 \\ %& 28.04 & 5.714 \\
\quad +CA-Redist (a) & \cellcolor{lightgray!35} 26.15 & \cellcolor{lightgray!35} 4.618 \\% & 15.70 \\ %& 26.43 & 4.719 \\
\bottomrule
    \end{tabular}
    \caption{FID and KID w.r.t. COCO2017 validation images.
    }
    \label{tab:fidkid}
\end{table}

\paragraph{Comparison of attention control methods:}

For \textbf{CAC}-style control, we observe that it does help resolve ambiguity, but not very consistently across different seeds.
In the first column, for example, the assignment completely failed.
We also see that it does not resolve the issue of improbable assignments, leaving the images largely the same as plain ControlNet for the rabbit example.

The other methods appear to provide satisfactory degree of control over object assignment in most cases.
However, as we can see in the last two columns of Fig.~\ref{fig:qualitative}, for the prompt ``\textit{an apricot, a pumpkin, and an orange}'',
\textbf{DenseDiffusion} and \textbf{eDiff-I} suffer from the aforementioned control scheduling problems, where objects of similar color and shape are not assigned correctly. 
Both
%In this case, in Fig.~\ref{fig:qualitative}, these methods 
generate two or three pumpkins, ignoring other described objects.
We observed similar behavior with other test cases.

\afk{In contrast}, CA-Redist adheres to localized descriptions better than DenseDiffusion and eDiff-I in more challenging control scenarios (objects of similar shape and color) while maintaining image quality.

\paragraph{Ablation:} The bottom three rows of \cref{fig:qualitative} show an ablation of CA-Redist, which shows it is still effective when only $W_m$ is non-zero (CA-Redist (m)) or only $W_a$ is non-zero (CA-Redist (a)).
When both are zero (CA-Redist (none)), the images don't always satisfy all region descriptions.
This shows that some form of attention boosting is still necessary.

\subsection{Quantitative study}
\label{sec:empirical}

Automatic evaluation for this task is rather challenging because of the inherent difficulty of evaluating image quality and faithfulness to a localized description. 
Moreover, there is a lack of standardized open-sourced dataset and evaluation methodology.
We chose to perform evaluation using a challenging dataset of simple scenes with multiple objects (\textsc{SimpleScenes}), and to verify the distributional properties %of the modified ControlNet 
using FID and KID on COCO2017.
%
%More details about the quantitative evaluation are provided in \cref{supp:eval}.

\paragraph{\textsc{SimpleScenes}} In this work, we are mainly interested in measuring (1) image quality to detect image degradation and (2) faithfulness to localized descriptions.\footnote{ControlNet adheres to the segmentation map conditioning very well in all cases so we do not evaluate this.}
Since COCO images frequently contain objects with overlapping bounding boxes, these could pollute the CLIP-based metrics we use to measure conformity to localized descriptions.

\noindent \textbf{Dataset:} For this reason, and to focus on more challenging cases, we create the \textsc{SimpleScenes} dataset consisting for a large part of segmentation maps for objects with less overlapping bounding boxes.
The dataset consists of 124 examples, each containing 3-4 objects and randomized descriptions, which proved to be challenging for the tested methods.
The dataset is described in more detail in \cref{sec:supp:simplescenes} in supplementary material.

\begin{figure*}[!t]
    \centering
    \noindent
    \includegraphics[width=0.8\textwidth]{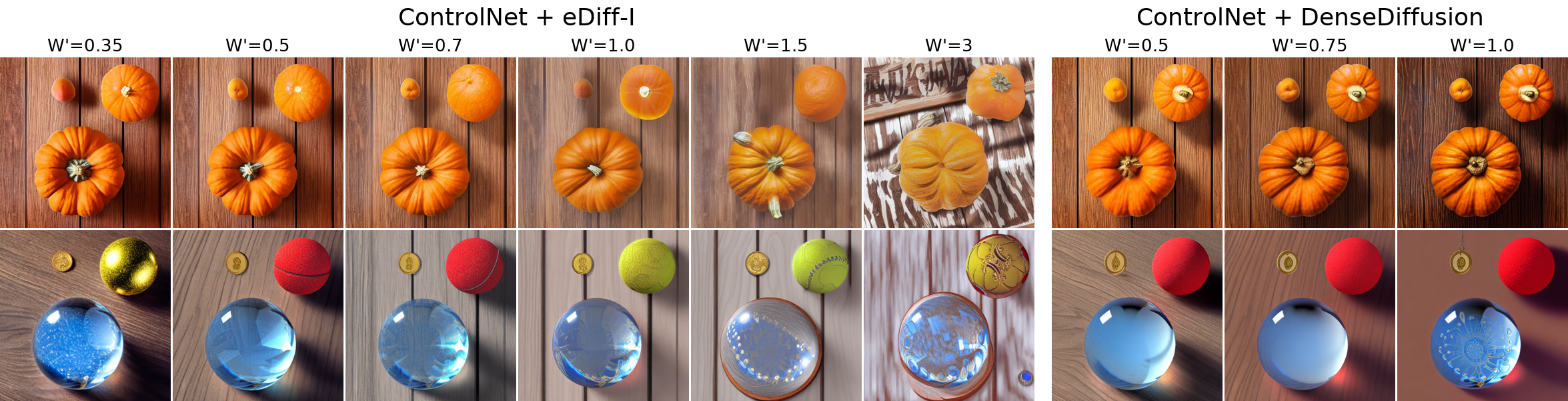}
    \caption{Image quality with increasing control strength for eDiff-I and DenseDiffusion cross-attention control with ControlNet.}
    % \captionof{figure}{Image quality with increasing control strength for eDiff-I and DenseDiffusion cross-attention control with ControlNet.}
    \label{fig:tradeoff}
\end{figure*}

\noindent \textbf{Metrics:} We use the following evaluation methodology for this dataset.
For measuring general image quality, we use reference-free\footnote{Since we don't have reference images.} image quality assessment methods, such as BRISQUE~\cite{brisque} and MANIQA~\cite{maniqa}, as well as the LAION Aesthetics Score predictor~\cite{laion}\footnote{We expect poor image quality to be inversely proportional to image aesthetics scores.}.
For measuring conformity to localized description, we use the Localized CLIP Logits and Probabilities, which are computed as follows:
For every object mask, we crop the image to contain only the masked region of the generated image, and use CLIP to compute text-image similarities between all the localized phrases (e.g. "blue crystal ball'') and all the cropped object images. 
The reported CLIP Logits correlate linearly with the similarities while the reported CLIP Probabilities \afk{result from} normalizing the logits over all the objects in the image.

\noindent \textbf{Results:} %From these 
\afk{The} numbers reported in Table~\ref{tab:quanteval} %, it appears 
\afk{demonstrate}
that CA-Redist does not suffer \afk{from} image quality loss, with all three image quality metrics being on par with plain ControlNet.
However, %from 
the Local CLIP metrics  %it appears 
\afk{indicate}
that %the 
CA-Redist %methods are 
\afk{is}
superior when it comes to conformity to localized descriptions.

\noindent \textbf{Ablation:}
From the three ablation settings (CA-Redist variants (a), (m) and (none)), it appears that moderately boosting attention does not result in measurable image quality decrease.
However, no attention boost (CA-Redist (none)) results in slightly lower faithfulness to the localized prompt (as indicated by the localized CLIP metrics), which confirms our observations from the qualitative study.

\paragraph{\textsc{COCO2017}} In addition, in Table~\ref{tab:fidkid}, we report FID~\cite{fid} and KID~\cite{kid} scores between 5000 samples generated using the segmentation maps from the COCO2017 validation set and the corresponding 5000 real images.
There we can see that even though ControlNet has worse FID and KID than GLIGEN, the addition of cross-attention control does not result in quality loss measurable by these metrics.
Note that GLIGEN and DenseDiffusion don't always adhere to the masks as closely as ControlNet since GLIGEN only uses bounding boxes instead of segmentation maps as input, and DenseDiffusion uses self-attention control.
Both don't follow the input segmentation maps as closely as ControlNet.
%, which may complicate the interpretation of the scores in \cref{tab:fidkid}.

\begin{figure*}[!t]
    \centering
    \includegraphics[width=0.8\textwidth]{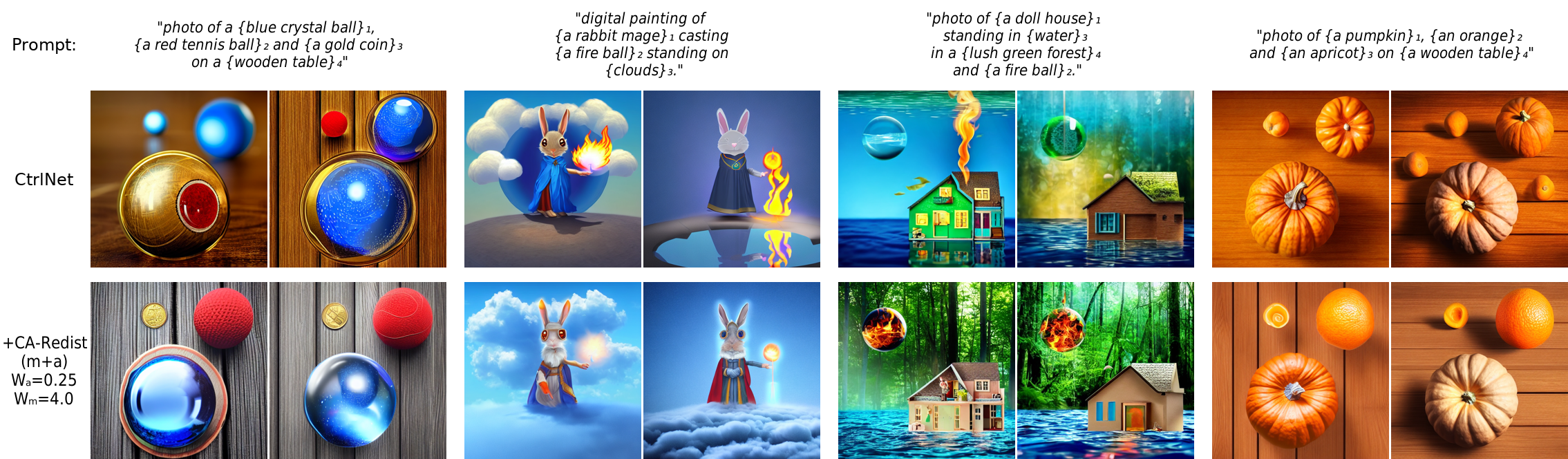}
    \caption{A qualitative comparison between sketch-based ControlNet with and without CA-Redist.
    See Supplement~\ref{sec:supp:scribbles} for a full comparison.
    }
    \label{fig:scribblecond}
\end{figure*} 

\subsection{Image quality vs control strength}
\label{sec:results:tradeoff}
In Fig.~\ref{fig:tradeoff} are shown some images generated using eDiff-I-based and DenseDiffusion-based attention control.
We can observe that at lower attention control strengths (controlled by the inference hyper-parameter $W'$), eDiff-I does not correctly follow localized descriptions.
Fidelity increases with increasing $W'$, however, it comes at a cost to image quality, which is particularly noticeably for higher values.
For DenseDiffusion, increasing $W'$ to its maximum value of $1.0$ does not improve fidelity to localized descriptions but also introduces more subtle image quality changes.

\section{Additional Analysis}
\label{sec:moreanal}
In this section, we provide additional analysis, focusing on (1) using other types of conditioning, (2) conditioning on more complex shapes and (3) a small ablation where we disable ControlNet.

\paragraph{Other types of layout conditioning:}
\label{sec:moreanal:othercond}
We performed our main investigation using segmentation map inputs, using the corresponding version of ControlNet.
Here, we show that the developed methods can also be used in combination with other types of conditioning.
In Fig.~\ref{fig:scribblecond}, we show that CA-Redist also performs well with ControlNet trained for sketch conditioning (scribbles).
A full qualitative comparison is provided in Supplement~\ref{sec:supp:scribbles}.

\begin{figure}
    \centering
    \includegraphics[width=\linewidth]{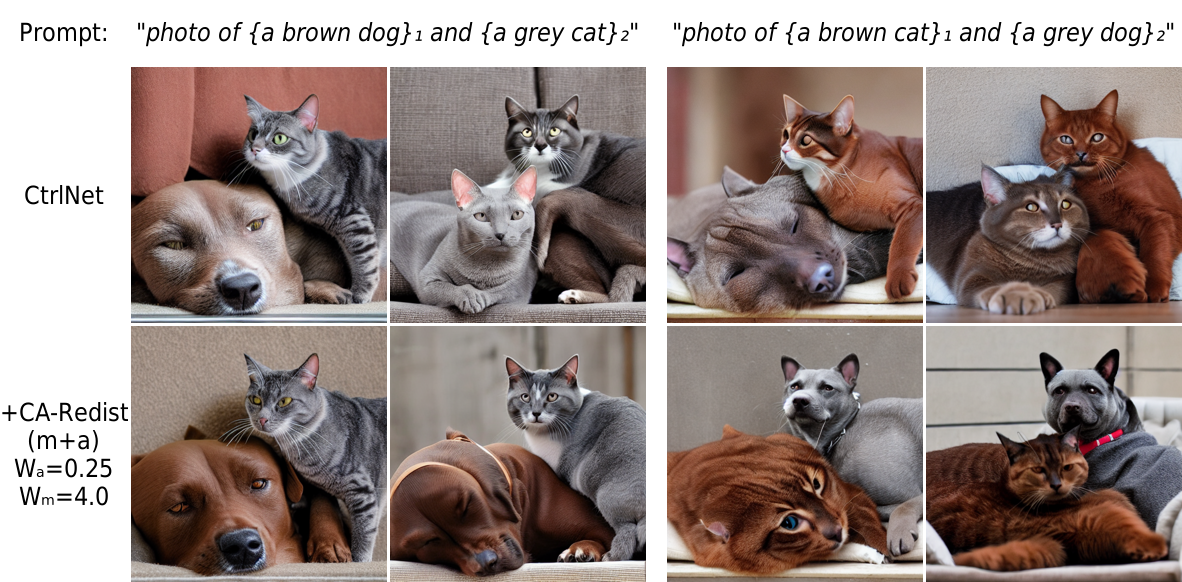}
    \caption{A qualitative comparison with complex shapes using an example from COCO.
    Region 1 is the bottom-left region and region 2 is the region on the right (see Fig.~\ref{fig:layouts} for exact layout specifications). 
    The figure shows that CA-Redist is able to assign the correct object identity. %, swapping cats and dogs when given the same conditioning input to ControlNet. 
    For a more elaborate comparison using complex shapes, see Supplement~\ref{sec:supp:complex}.
    }
    \label{fig:complexshapes}
\end{figure}

\paragraph{Using more complex shapes:}
We also performed an analysis to show that CA-Redist also works with more complex shapes, as illustrated in Fig.~\ref{fig:complexshapes}.
This confirms the general applicability of the investigated methods.

\begin{figure}
    \centering
    \includegraphics[width=\linewidth]{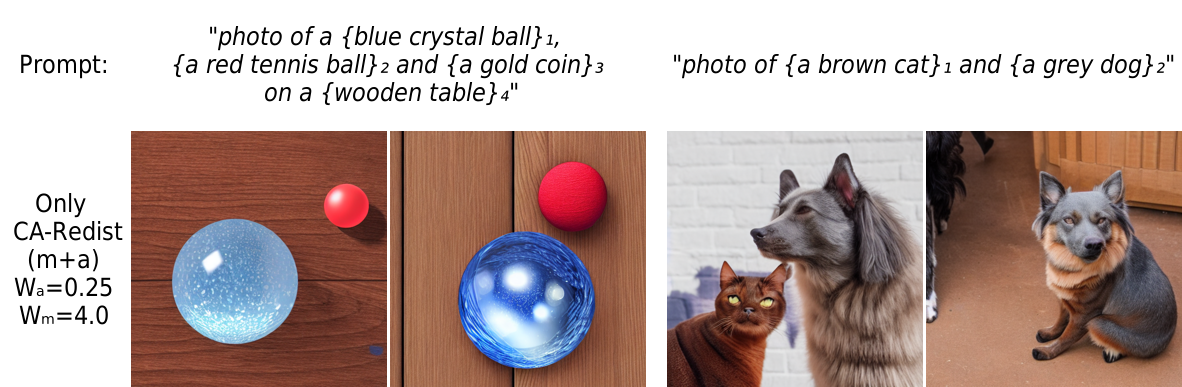}
    \caption{Examples of using only CA-Redist (no ControlNet).
    }
    \label{fig:nocontrol}
\end{figure}

\paragraph{Disabling ControlNet:} When ControlNet is disabled, cross-attention control is insufficient to generate all objects and ensuring the specified object boundaries are respected. This is illustrated in Fig.~\ref{fig:nocontrol}.

% TODO: how CA-Redist behaves

% TODO: is this just about when to put threshold?

% TODO: add some table with numbers

\section{Related Work}
Since the early works on controllable image synthesis~\cite{johnson2006semantic,chen2009sketch2photo}, the emergence of neural-network based generative models opened new frontiers for this task, especially with methods like ControlNet~\cite{controlnet} and GLIGEN~\cite{gligen}.
Several works have addressed tasks similar to ours with GANs~\cite{park2019semantic,sushko2020you,zhao2019image,zhu2020sean,sun2019image,li2021image}.
However, these works either do not use descriptions or are limited to a restricted set of object classes.

More recently, several works have proposed methods~\cite{layoutdiffusion,scenecomposer,spatext,multidiffusion,densediffusion,cac,ediffi,chen2023training,phung2023grounded,rnb,zestguide,mao2023training} for image synthesis with localized descriptions and regions specified by masks or bounding boxes. 
SceneComposer~\cite{scenecomposer} and SpaText~\cite{spatext} modify and train the diffusion model to condition on localized prompts.
Proper comparison with these methods was not possible because no code or data has been released by the authors at the time of this writing and they require a high computational budget for training.
Several training-free methods for this task have been proposed~\cite{multidiffusion,densediffusion,cac,ediffi,chen2023training,phung2023grounded,rnb,zestguide,mao2023training}, many of which rely on manipulating cross-attention (and self-attention) in some way.
Multidiffusion~\cite{multidiffusion} proposes a method that requires running a separate denoising process for every separate region, thus introducing significant overhead.

Cross-attention modification have also been a useful tool in an array of other approaches involving diffusion models, such as improving image quality and conformity to the prompt~\cite{attendandexcite,agarwal2023star} and other tasks~\cite{epstein2023diffusion,hertz2022prompt,structurediffusion,masactrl}.

% Different previous works perform different evaluations. For eDiff-I~\cite{ediffi}, the authors did not provide any quantitative evaluation.

% The authors of SceneComposer~\cite{scenecomposer} perform evaluation TODO

% The authors of CAC~\cite{cac} provide an extensive evaluation where they look at fidelity, controllability and inference time.
% Kernel Inception Distance (KID)~\cite{kid} is used for fidelity and measures how closely the generated images look to the images from a reference dataset (COCO-2017).
% Controllability is evaluated by how well bounding boxes (extracted using YOLOv8~\cite{yolov8}) or segmentation maps (extracted using SSA~\cite{ssa}) match the ground truth annotations that were used during generation.
% In addition, compositional generation was evaluated using the CC-500 dataset, which consists of prompts describing an object and a color.

\section{Conclusion}
\label{sec:conclusion}
{The analysis performed in this work indicates that ControlNet is able to interpret region descriptions when mask shapes are un-ambiguous.
However, when faced with similarly shaped masks, it no longer has sufficient information to correctly interpret the prompt. \afk{As we demonstrate,}
this can be solved by integrating cross-attention control. \afk{Such control techniques augment}
segmentation-based ControlNet's ability to very closely follow region masks %is extended 
with the ability to control the contents of each of these regions more precisely.
We found, however, that some design choices are crucial in more ambiguous conditions, such as when generating multiple different objects of similar shape and color.
To cover these cases better, it is important to prevent cross-attention from attending to tokens from irrelevant region descriptions throughout the \textit{entire} generation process.
\afk{Taking these considerations into account,}
%Finally, 
we develop a novel cross-attention control method that  %takes into account these considerations. 
\afk{shows superior generation results in our qualitative and quantitative analysis. For the latter we created a small but challenging data set, which can serve as a testbed for future work.}
Our code is open-sourced on Github\footnote{\url{https://github.com/lukovnikov/ca-redist}}.

\clearpage

{
    \small
    \bibliographystyle{ieeenat_fullname}
    \bibliography{main}
}

\clearpage
\setcounter{page}{1}
\maketitlesupplementary

\section{Experimental Setup}
\label{sec:supp:setup}
In this section, we elaborate more on the experimental setup used in this work that did not fit in the main paper due to space limitations.
The rest of this section is structured as follows.
First, the \textsc{SimpleScenes} dataset that was used in the qualitative and the quantitative analysis is described (\cref{sec:supp:simplescenes}), then the general setup for both the qualitative and quantitative analysis \afk{is detailed} (\cref{sec:supp:qualitative}).
In \cref{sec:supp:quantitaive}, the experimental setup for the quantitative analysis is elaborated in more detail.
All our code and data will be made publicly available after the double-blind review process.

\subsection{SimpleScenes dataset}
\label{sec:supp:simplescenes}
Whereas several previous works \afk{were} evaluated using the COCO~\cite{coco} dataset, in this work, we chose to create a new dataset to explicitly test more challening generation conditions.
%Firstly, ControlNet was trained on COCO data, and thus might fit it better.
The main reason \afk{for this decision} is that the large masks in COCO are generally detailed, and their shape gives away what objects could be expected there, reducing the risk of ambiguous generation conditions.

We create a small dataset consisting of segmentation maps and localized descriptions that present the generator with more ambiguous cases that challenge the ability of the tested methods to correctly assign region descriptions to regions in the generated image.
The first problematic case is when there is shape ambiguity, where multiple regions of similar shapes (e.g., two circles) are present but have to be assigned different objects (e.g., crystal ball and gold coin).
A more problematic case is the presence of shape and color ambiguity (e.g., an orange and a pumpkin).
In the latter case, partially noisy versions of an image, like the ones observed in the middle of the denoising process, can still be ambiguous and prevent a grounding method from assigning the right textures and other details to the objects.

To generate the dataset, we first manually create different segmentation maps.
Then, for every segmentation map, we define one or more sets of objects descriptions.
The sets of objects descriptions and the segmentation maps they are used with are provided in \cref{fig:supp:simplescenes:spec}.
Finally, for each segmentation map and set of object descriptions, we generate a set of examples. 
For every example, we randomly select a subset of objects and assign to every shape in the segmentation map a description randomly from the selected objects. A localized prompt with the sampled object assignment is then composed.
In addition to ambiguous maps, we also include an example that was used in eDiff-I~\cite{ediffi}, and another less ambiguous example.
In total, we have 124 unique examples, most of which have proven to be challenging.

\begin{table*}[!htb]
    \centering
    \begin{tabular}{p{0.1\linewidth} p{0.8\linewidth} p{0.1\linewidth}}
        Seg. map & Prompt or Template & \# examples \\
        \midrule
        \parbox[c]{\linewidth}{\includegraphics[trim={0 0 1536px 0},clip, width=\linewidth]{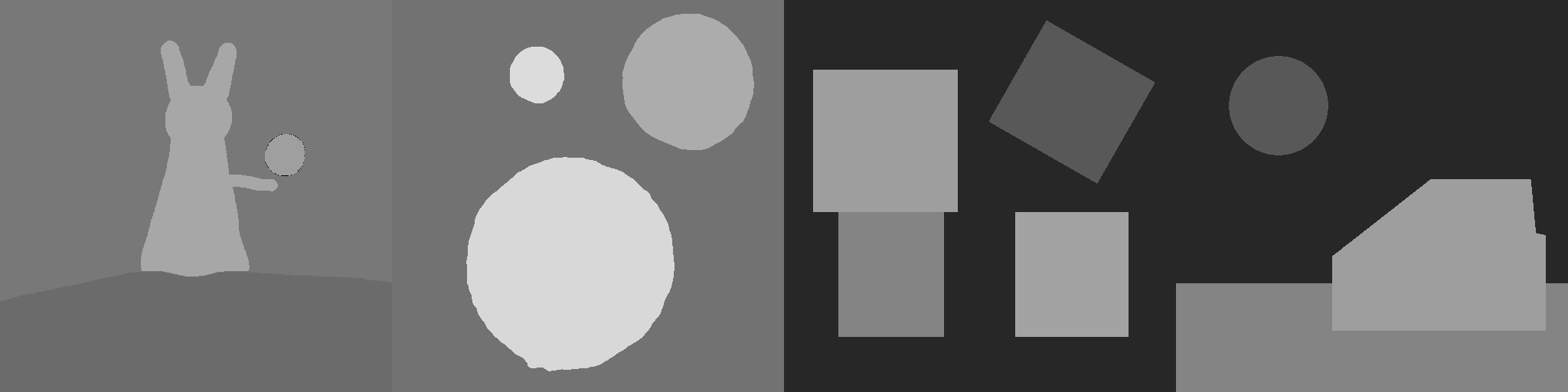}}     & {\small ``a digital art of a rabbit mage standing on clouds casting a fire ball. Background is a blue sky.'' } & 2 \\ ~\\
        \parbox[c]{\linewidth}{\includegraphics[trim={512px 0 1024px 0},clip, width=\linewidth]{images/allmasks.png}} & \parbox[c]{\linewidth}{{\small ``a highly detailed photorealistic image of a gold coin, a blue crystal ball and a red tennis ball on a wooden table.''}} & 2 \\~\\
        \parbox[c]{\linewidth}{\includegraphics[trim={512px 0 1024px 0},clip, width=\linewidth]{images/allmasks.png}} & \parbox[c]{\linewidth}{{\small
        A = \{"an orange", "a pumpkin", "an apricot", "a persimmon" \}  \\
        Template = ``a highly detailed photorealistic image of \{A\}, \{A\} and \{A\} on a wooden table''
        }} & 20 \\~\\
        \parbox[c]{\linewidth}{\includegraphics[trim={512px 0 1024px 0},clip, width=\linewidth]{images/allmasks.png}} & \parbox[c]{\linewidth}{{\small
        C = \{"red", "blue", "yellow"\}\\
        A = \{"a \{C\} crystal ball", "a \{C\} tennis ball", "a \{C\} ping pong ball" \}  \\
        B = \{"a wooden table", "grass"\}\\
        Template = ``a highly detailed photorealistic image of \{A\}, \{A\} and \{A\} on \{B\}''
        }}
        & 20 \\~\\
        \parbox[c]{\linewidth}{\includegraphics[trim={512px 0 1024px 0},clip, width=\linewidth]{images/allmasks.png}} & \parbox[c]{\linewidth}{{\small
        C = \{"red", "blue", "green", "pink", "white", "yellow"\} \\
        B = \{"a wooden table", "grass"\}\\
        Template = ``a highly detailed photorealistic image of a \{C\} ball, a \{C\} ball and a \{C\} ball on \{B\}''
        }} & 20 \\~\\
        \parbox[c]{\linewidth}{\includegraphics[trim={512px 0 1024px 0},clip, width=\linewidth]{images/allmasks.png}} & \parbox[c]{\linewidth}{{\small
        A = \{"an orange", "a red apple", "a green apple", "a watermelon", "a kiwi", "a chestnut", "a walnut", "a peach", "a grape"\} \\
        B = \{"a wooden table", "grass"\}\\
        Template = ``a highly detailed photorealistic image of \{A\}, \{A\} and \{A\} on \{B\}''
        }} & 20 \\~\\
        \parbox[c]{\linewidth}{\includegraphics[trim={1024px 0 512px 0},clip, width=\linewidth]{images/allmasks.png}} & \parbox[c]{\linewidth}{{\small
        A = \{"an old-school tv", "a cardboard box", "a square watermelon", "a concrete block", "spongebob", 
           "a yellow book", "a washing machine", "a gopro camera"\} \\
        B = \{"a lush forest", "a brick wall", "a living room"\}\\
        Template = ``a photo of \{A\} on top of \{A\}, and to the right \{A\} falling on \{A\}. Background is \{B\}.''
        }} & 20 \\~\\
        \parbox[c]{\linewidth}{\includegraphics[trim={1536px 0 0 0},clip, width=\linewidth]{images/allmasks.png}}     & \parbox[c]{\linewidth}{{\small
        A = \{"a doll house", "a container ship", "an old shack", "a car"\} \\
        B = \{"in the desert", "floating in the sea", "standing on the ground", "standing on grass"\}\\
        D = \{("a dark starry night sky", \{"the moon", "a blood red moon", "a hot air balloon"\}), ("a blue sky", \{"the sun", "the moon", "a hot air balloon", "the death star"\}), ("a lush green forest", \{"a magical fire orb", "a red balloon"\})\} \\
        Template = ``a photo of \{A\} \{B\} Background is \{$\text{D}_{1}$\} with \{$\text{D}_{2}$\}.''
        }} & 20 \\
    \end{tabular}
    \caption{Prompts and templates used to generate examples for the \textsc{SimpleScenes} dataset.
    Templates are provided together with sets of objects that can be used to fill placeholders with the same set name in the template. $\text{D}_1$ and $\text{D}_2$ in the last template means that we sample one of the elements of D (which are two-tuples), and take the first element of the tuple as $\text{D}_1$ and randomly select one value from the second element of the tuple as $\text{D}_2$.
    Examples are generated by randomly sampling values for placeholders in the templates from the corresponding sets of object descriptions.
    }
    \label{fig:supp:simplescenes:spec}
\end{table*}

\subsection{Experimental setup}
\label{sec:supp:expsetup}
The settings described below have been used both in the qualitative (\cref{sec:qualitative}) and quantitative (\cref{sec:empirical}) analysis. Please see \cref{sec:supp:quantitaive} for more details on quantitative analysis.

ControlNet~\cite{controlnet} v1.1 trained on ADE20k~\cite{ade40k} and COCO~\cite{coco} segmentation maps was used for the experiments.
See \cref{sec:supp:controlnet} for a short description of ControlNet and example usage.
We fine-tuned ControlNet using COCO 2017 Panoptic training data while randomizing the colors of different objects to match the random colors used during inference.
While fine-tuning, only the \texttt{\small input\_hint\_block} of the control model was trained.
We also experimented with fine-tuning (1) the entire control model, (2) first control model layer from the copied portion of the original U-Net, (3) only the first three layers of the input hint block. 
None of these settings gave better results than the one we report in the paper in our initial testing.
Fine-tuning of ControlNet was done using the Adam~\cite{adam} optimizer with a learning rate of $10^{-5}$ for 50k training steps, which takes approximately one day on a single GPU.

We used the DDIM sampler with 50 steps to generate images for all the models compared in \cref{fig:qualitative} in \cref{sec:qualitative}.
Besides multiplicative and additive attention boost strengths ($W_m$ and $W_a$, respectively), CA-Redist also allows to control the attention boost schedule using the threshold timestep $T_{\text{thr}}$ and softness $R$, as elaborated in \cref{sec:method}.
The threshold timestep $T_{\text{thr}}$ for all the CA-Redist results reported in \cref{fig:qualitative} is set to $T$.
The softness for CA-Redist (m+a) and CA-Redist (m) is set to 0.8 and for CA-Redist (a) to 0.6.
These values mean that the attention boost is only active in the first 40\% or 30\% of the decoding process, with a strong decay, similarly to the integrated eDiff-I and DenseDiffusion.
Note that the attention redistribution of CA-Redist still remains active throughout the entire decoding process and is not affected by the attention boost schedule.

For the GLIGEN and DenseDiffusion (original) experiments, we also used 50 DDIM steps, keeping all values default, unless otherwise specified in the figures.

Stable Diffusion v1.5 was used for ControlNet (since it was pre-trained with it) and the original DenseDiffusion.
For GLIGEN, we used the pre-trained checkpoint provided in the Huggingface Diffusers library. 
This checkpoint uses Stable Diffusion v1.4.

\subsection{Quantitative analysis}
\label{sec:supp:quantitaive}
For the quantitative analysis, we would like to measure both (1) image quality and (2) faithfulness of the generated image to the localized descriptions.
Since we don't have reference images, usual metrics like FID are not applicable.
For this reason, we use reference-free image quality metrics (BRISQUE and MANIQA), and the pre-trained LAION Aesthetics predictor.
For BRISQUE and MANIQA, we use their implementations from PyIQA~\footnote{\url{https://github.com/chaofengc/IQA-PyTorch}}.
For the LAION Aesthetics predictor, we use the implementation provided by LAION~\footnote{\url{https://github.com/LAION-AI/aesthetic-predictor/}}.

For assessing faithfulness of the generated images to the localized descriptions, we implement localized CLIP logits and probability-based metrics.
In most examples in our dataset, most objects have non-overlapping bounding boxes, which is useful for evaluation with CLIP.
For every object mask, we crop the image to contain only the masked region of the generated image, and use CLIP to compute text-image similarities between all the localized phrases (e.g., "blue crystal ball'') and all the cropped object images. 
The reported CLIP logits correlate linearly with the dot products between the image and text vectors while the reported CLIP probabilities result from normalizing the logits over all the object descriptions in the image. 
% \af{[Check: DDo we really want to write logits and probabilties after CLIP with upper case letters? Is this consistently done?]}
The reported numbers are the logits and probabilities between the region in the image and the \textit{correct} region description.
While the probabilities are derived solely from the logits, and strongly correlate, we believe it's better to report both, since the logits are un-normalized and only describe the affinity of an image region with a given region description, regardless of other region descriptions for the same image.
On the other hand, normalizing the logits %presents
\afk{results in}
a metric that takes into account all available object descriptions.
If some region in the image is ambiguous, it could have high affinity with two or more region descriptions.
This will not necessarily be reflected in Local CLIP logits, but would result in a much lower Local CLIP probability.

Finally, we would like to elaborate on how exactly the numbers in \cref{tab:quanteval} were computed.
For every example in the dataset, we generated five images from different seeds and computed all metrics for every generated image.
As specified in the table description, the reported numbers follow the format $\mathrm{MEAN}^{\pm \mathrm{STD}} (\mathrm{BEST})$.
To obtain the \textbf{means} and \textbf{standard deviations} reported in \cref{tab:quanteval}, for every metric, we first take its average \textit{over examples}, giving us one value for every seed.
The reported means are computed by then taking the average \textit{over seeds}. 
Also the standard deviations are computed over seeds in order to give a sense of the spread of the outcomes of multiple experiments.
For the \textbf{best} number, however, we simulate the case where per example, five candidates are generated using different seeds and the best one is picked by the user.
Since we can't exactly predict which image would be picked by the user, we simply compute the best scores by taking the argmax (or argmin for metrics where lower is better) over the seeds for every example, and subsequently averaging over all the examples.

% \vspace{7em}
\section{A note on \cref{eq:cosineschedule}}
\label{sec:supp:eq15}
% For a more elaborate version of \cref{eq:cosineschedule} in the main text, please consider the following:
% \begin{align}
%     W'' &= \begin{cases}
%         1 & \text{if $t \geq T_{\mathrm{s}}$} \\
%         \frac{1}{2} + \frac{1}{2} \sin (\pi \cdot \frac{t - T_{\mathrm{thr}}}{T_{\mathrm{s}} -  T_{\mathrm{e}}})   & \text{if $T_{\mathrm{s}} > t > T_{\mathrm{e}}$} \\
%         0 & \text{if $T_{\mathrm{e}}$} \geq t  \\
%     \end{cases} \\
%     T_{\mathrm{s}} &= T_{\mathrm{thr}} + T \cdot R/2\\
%     T_{\mathrm{e}} &= T_{\mathrm{thr}} - T \cdot R/2
% \end{align}
% The weight $W''$ specifies the schedule of attention boost in CA-Redist and depends on the current denoising step $t$.
% This schedule is controlled by the threshold step $T_{\mathrm{thr}} \in [1 \enleadertwodots T ]$ and the threshold softness $R \in [0, 1]$.
In \cref{eq:cosineschedule}, $t$ starts from $T$ so attention boost is active more in the initial stages of denoising with a value between zero and one and gradually decays to zero as denoising progresses.

When $T_{\mathrm{thr}} = T$, we can simplify the schedule as follows:
% \begin{align}
%     W'' &= \begin{cases}
%         \frac{1}{2} + \frac{1}{2} \sin (\pi \cdot \frac{t - T}{R\cdot T})   & \text{if $t > T_{\mathrm{e}}$} \\
%         0 & \text{if $T_{\mathrm{e}}$} \geq t  \\
%     \end{cases} \\
%     T_{\mathrm{e}} &= T \cdot (1 - R/2)
% \end{align}
%
\begin{align}
    W'' &= \begin{cases}
        \frac{1}{2} + \frac{1}{2} \sin (\pi \cdot \frac{t - T}{R\cdot T})   & \text{if $t > T \cdot (1 - R/2)$} \\
        0 & \text{otherwise}  \\
    \end{cases}
\end{align}

This schedule for $R=0.4$ is illustrated in \cref{fig:schedule}.
\begin{figure}[!h]
    \centering
    \includegraphics[width=0.75\linewidth]{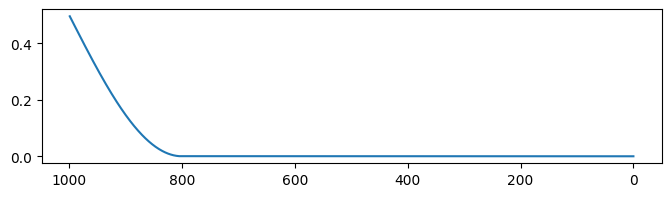}
    \caption{CA-Redist schedule $W''$ if $T_{\mathrm{thr}} = T$ and $R=0.4$.}
    \label{fig:schedule}
\end{figure}

% \noindent\textbf{Note on \cref{eq:dd2}:} As used in \cref{eq:dd2}, $t$ goes from $0$ (only noise) to $T$ (no noise) during denoising, in contrast to \cref{eq:cosineschedule}, for which $t$ was defined exactly the opposite, going from $T$ (only noise) to $0$ (no noise).
% If $t$ is used in \cref{eq:dd2} as defined for \cref{eq:cosineschedule}, then \cref{eq:dd2} should be written as:
% {\small
% \begin{align}
%     W =& ~ W' \cdot \big(  \dfrac{t}{T}  \big) ^5 \cdot (1 - S)
%     \odot ( \mathbf{B}_{f_{\mathrm{RT}}(n)} \odot \mathbf{M}_{\mathrm{+}}
%    - (1-\mathbf{B}_{f_{\mathrm{RT}}(n)}) \odot  \mathbf{M}_{\mathrm{-}} ) \nonumber
% \end{align}
% }

% \clearpage
% \vspace{7em}

\section{Illustration of attention redistribution}
\label{sec:supp:redist}
In this section, we provide an illustration of how the attention redistribution of CA-Redist (described in \cref{eq:caredist1,eq:caredist2,eq:caredist3,eq:caredist4}) works.

We illustrate this with a hypothetical example where the prompt consists of 12 tokens and some values for $\mathbf{Q} \mathbf{K}^T$ are given.
For example, consider the prompt ``\textit{A very realistic painting of \{dogs\} and \{cats\} and \{guinea pigs\} playing.}''.

\cref{fig:c_orig,fig:b_frt,fig:b_R,fig:c_local,fig:c_global,fig:c_new} shows the attention distributions over tokens for one position in the latent feature vectors of the U-Net.
In the context of our example, suppose that the chosen element belongs to the region where guinea pigs should be drawn.
Thus, tokens at positions 10 and 11 are relevant while 6 (dogs) and 8 (cats) are not.

% \noindent%
% \begin{minipage}{\textwidth}% to keep image and caption on one page
% \makebox[\linewidth]{%        to center the image
%     \centering
%   \includegraphics[width=0.95\textwidth]{images/qualitative5_full.png}}
% \captionof{figure}{Full-resolution version of \cref{fig:qualitative}.}\label{fig:qualitative:fullres}
% \end{minipage}

\noindent%
\begin{minipage}{\linewidth}% to keep image and caption on one page
\makebox[\linewidth]{%        to center the image
    \centering
    \includegraphics[width=\textwidth]{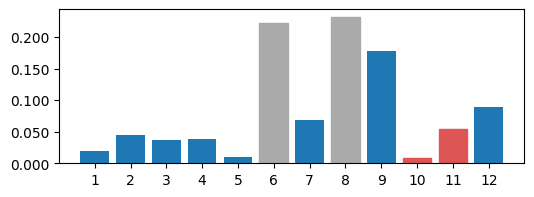}}
    \captionof{figure}{The original attention weights as computed in \cref{eq:attn}. 
    Blue bars indicate tokens not belonging to any region description.
    Grey and red bars indicate tokens belonging to any region description.
    Red bars indicate tokens belonging to the correct region description.
    }\label{fig:c_orig}
\end{minipage}

\noindent%
\begin{minipage}{\linewidth}% to keep image and caption on one page
\makebox[\linewidth]{%        to center the image
    \centering
    \includegraphics[width=\linewidth]{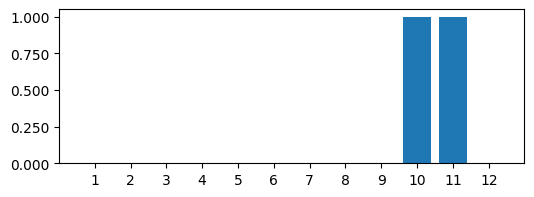}}
    \captionof{figure}{$\mathbf{B}_{f_{\mathrm{RT}}(n)}$.
    Only tokens at positions 10 and 11 belong to the relevant region description for the considered element in the feature vector in the U-Net.
    }
    \label{fig:b_frt}
\end{minipage}

\noindent%
\begin{minipage}{\linewidth}% to keep image and caption on one page
\makebox[\linewidth]{%        to center the image
    \centering
    \includegraphics[width=\linewidth]{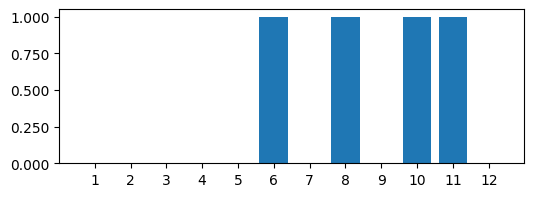}}
    \captionof{figure}{$\mathbf{B}_{R}$. 
    Tokens at positions 6, 8, 10, 11 belong to one of the region descriptions.
    }
    \label{fig:b_R}
\end{minipage}

\noindent%
\begin{minipage}{\linewidth}% to keep image and caption on one page
\makebox[\linewidth]{%        to center the image
    \centering
    \includegraphics[width=\linewidth]{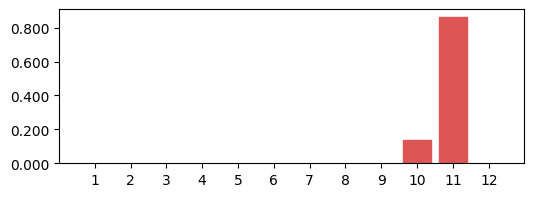}}
    \captionof{figure}{Attention weights as computed for $\mathbf{C}_{\mathrm{local}}$ in \cref{eq:caredist2}.}
    \label{fig:c_local}
\end{minipage}

\noindent%
\begin{minipage}{\linewidth}% to keep image and caption on one page
\makebox[\linewidth]{%        to center the image
    \centering
    \includegraphics[width=\linewidth]{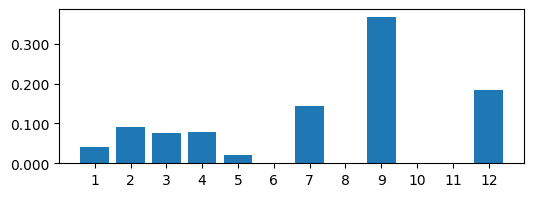}}
    \captionof{figure}{Attention weights as computed for $\mathbf{C}_{\mathrm{global}}$ in \cref{eq:caredist3}. Note that positions 6, 8, 10 and 11 have zero weight here as they belong to one of the region description.}
    \label{fig:c_global}
\end{minipage}

\noindent%
\begin{minipage}{\linewidth}% to keep image and caption on one page
\makebox[\linewidth]{%        to center the image
    \centering
    \includegraphics[width=\linewidth]{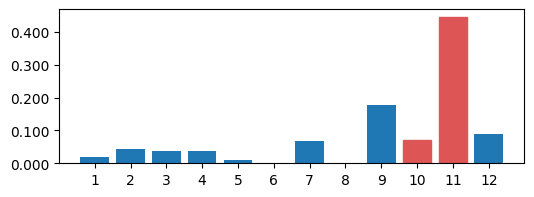}}
    \captionof{figure}{The new attention weights as computed for CA-Redist in \cref{eq:caredist1}. 
    Note that tokens at positions 10 and 11 now have the same total attention weight as all the region-specific tokens before redistribution.
    }
    \label{fig:c_new}
\end{minipage}

\noindent%
\begin{minipage}{\linewidth}% to keep image and caption on one page
\makebox[\linewidth]{%        to center the image
    \centering
    \includegraphics[width=\linewidth]{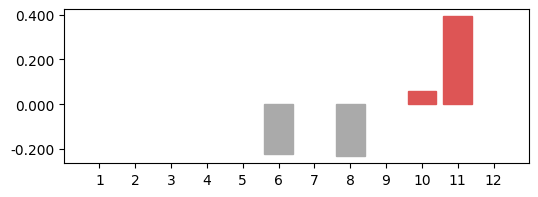}}
    \captionof{figure}{The difference between the new attention weights as computed for CA-Redist using \cref{eq:caredist1,eq:caredist2,eq:caredist3,eq:caredist4}) and the original attention weights as computed in \cref{eq:attn}.
    Note that attention to tokens not belonging to any region description is unchanged.
    }
    \label{fig:c_diff}
\end{minipage}

\clearpage
\section{Background: ControlNet}
\label{sec:supp:controlnet}
Here, we provide some more details about ControlNet for better understanding and refer the reader to the work of \citet{controlnet} for a more elaborate and detailed description.

In addition to the prompt $X$, ControlNet also expects an image $c_{\mathrm{img}}$ as input for the generation process.
In order to incorporate conditioning based on $c_{\mathrm{img}}$, first a control model is defined that copies the down-sampling and middle blocks of the latent diffusion model's U-Net (see \cref{fig:controlnet}). 
The control model also contains an additional block of convolutional layers (the \textit{input hint block}) that encodes the control signal $c_{\mathrm{img}}$ and is trained from scratch.
The features computed by the control model are added to the features computed by its sibling in the main U-Net before feeding them into the up-sampling blocks of the main U-Net. 

Adding features is done using \textit{zero-convolutions}, which control the degree to which each value of the added features contributes.
The zero-convolutions are $1 \times 1$ convolutions whose weights and biases are initialized to zero.
See \cref{fig:zeroconv} and \cref{fig:controlnet} for an illustration of zero-convolutions and how they are used in ControlNet.

\begin{figure}[b]
    \centering
    \includegraphics[width=\linewidth]{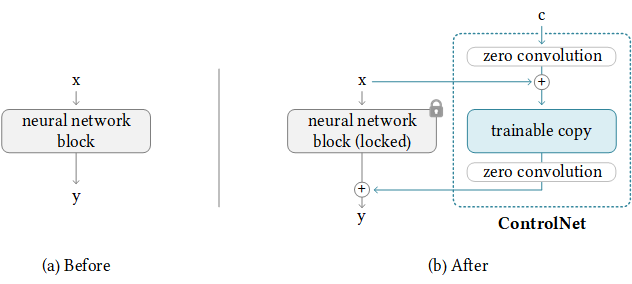}
    \caption{Before and after adapting a neural network block using zero-convolutions. The block is copied, while the original block is frozen. Only the zero-convolutions and the trainable copy are trained during ControlNet training.
    This image is from \cite{controlnet}.
    }
    \label{fig:zeroconv}
\end{figure}

\begin{figure}
    \centering
    \includegraphics[width=\linewidth]{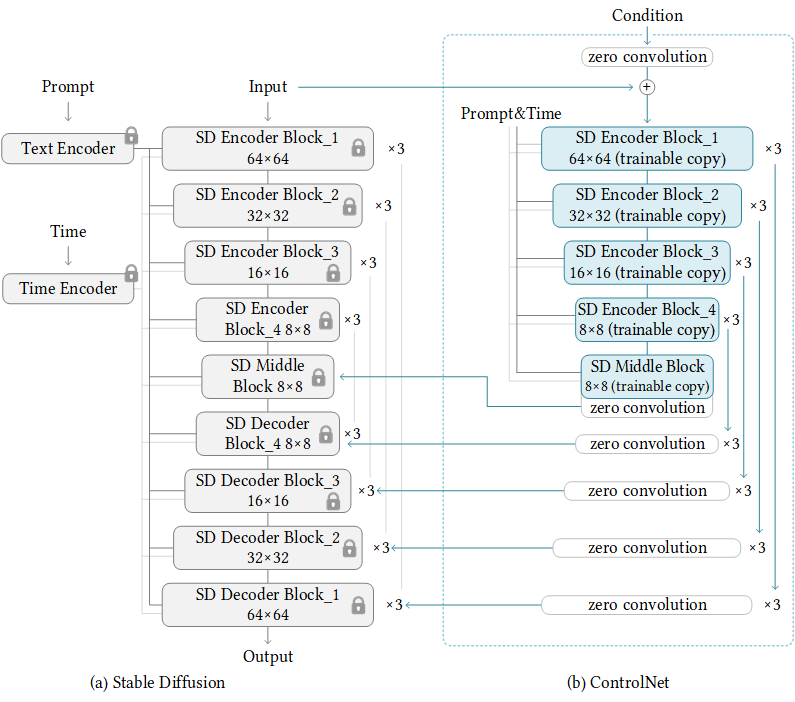}
    \caption{This figure shows the U-Net of Stable Diffusion and how ControlNet is applied there.
    The four input blocks (downsampling part) are copied while the original U-Net is frozen.
    Zero-convolutions control how the features from the copied input (down-sampling) blocks change the features fed into the middle and up-sampling blocks of the U-Net.
    Not shown in the figure is the network that processes the original conditioning image.
    This image is from \cite{controlnet}.
    }
    \label{fig:controlnet}
\end{figure}

An example of the use of segmentation-based ControlNet is given in \cref{fig:controlnet:segexample}.

\begin{figure}
    \centering
    \includegraphics[width=\linewidth]{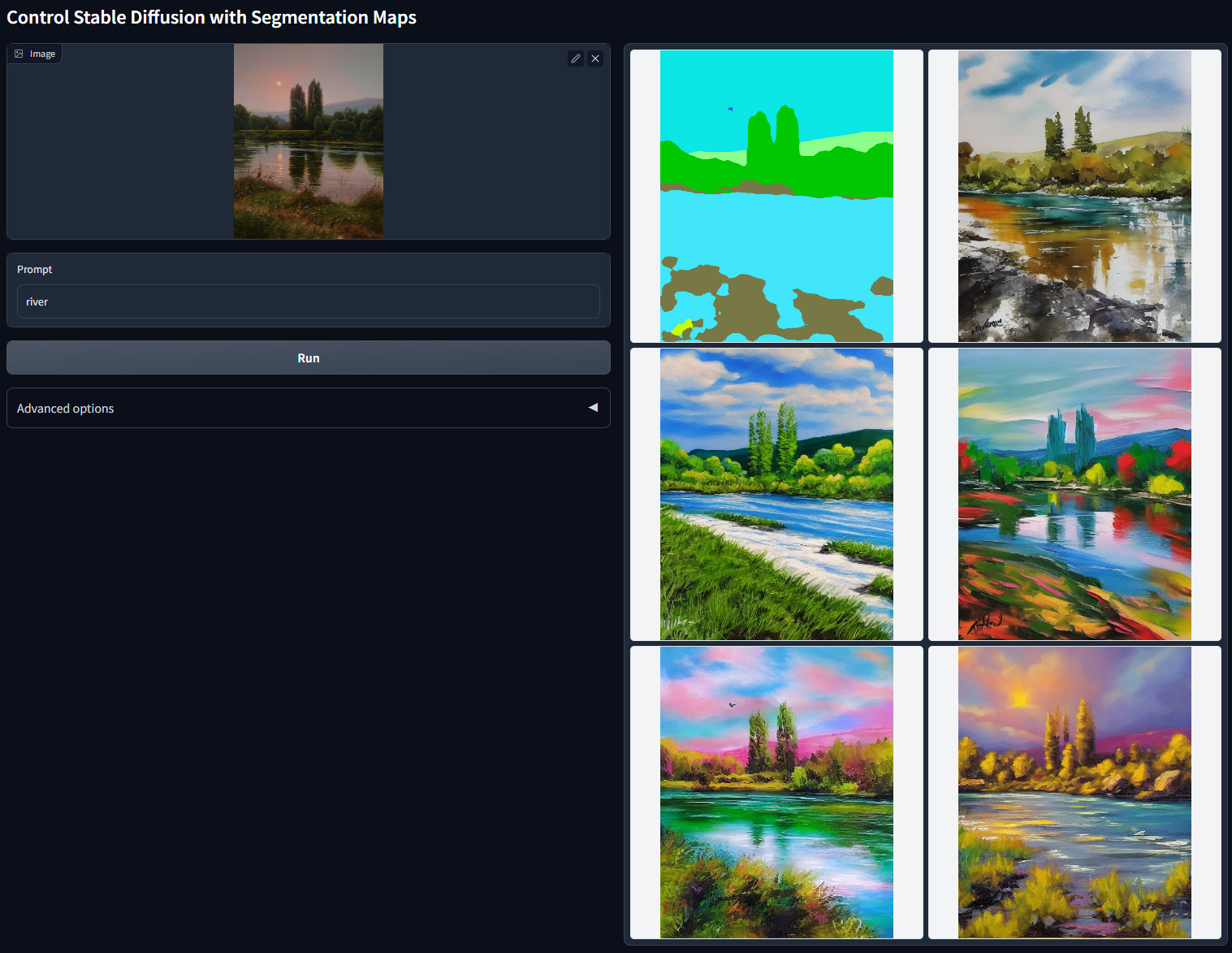}
    \caption{Example usage of segmentation-based ControlNet, as provided on \url{https://github.com/lllyasviel/ControlNet}.
    The segmentation map of the image on the left is extracted using a segmentation model.
    The extracted segmentation map is shown on the top left of the right part of the UI.
    Given the prompt and the segmentation map, Stable Diffusion with ControlNet is able to generate images with a composition as specified by the segmentation map.
    }
    \label{fig:controlnet:segexample}
\end{figure}

 \section{Qualitative analysis with Scribble conditioning}
\label{sec:supp:scribbles}

In Fig.~\ref{fig:qualitative_scribbles_full}, we show a qualitative analysis similar to Fig.~\ref{fig:qualitative} but for ControlNet conditioned on sketch input (scribbles). 
The input layouts are specified in Fig.~\ref{fig:scribblespecs}.
We can make observations similar to those in qualitative analysis in the main paper for segmentation map inputs (Fig.~\ref{fig:qualitative}).

\begin{figure}[!htb]
    \centering
    \includegraphics[width=0.7\linewidth]{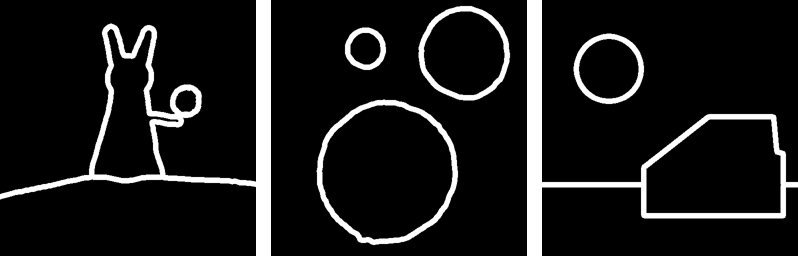}
    \caption{Scribbles used for Fig.~\ref{fig:qualitative_scribbles_full}.}
    \label{fig:scribblespecs}
\end{figure}

\begin{figure*}[!htb]
    \centering
    \includegraphics[width=0.99\textwidth]{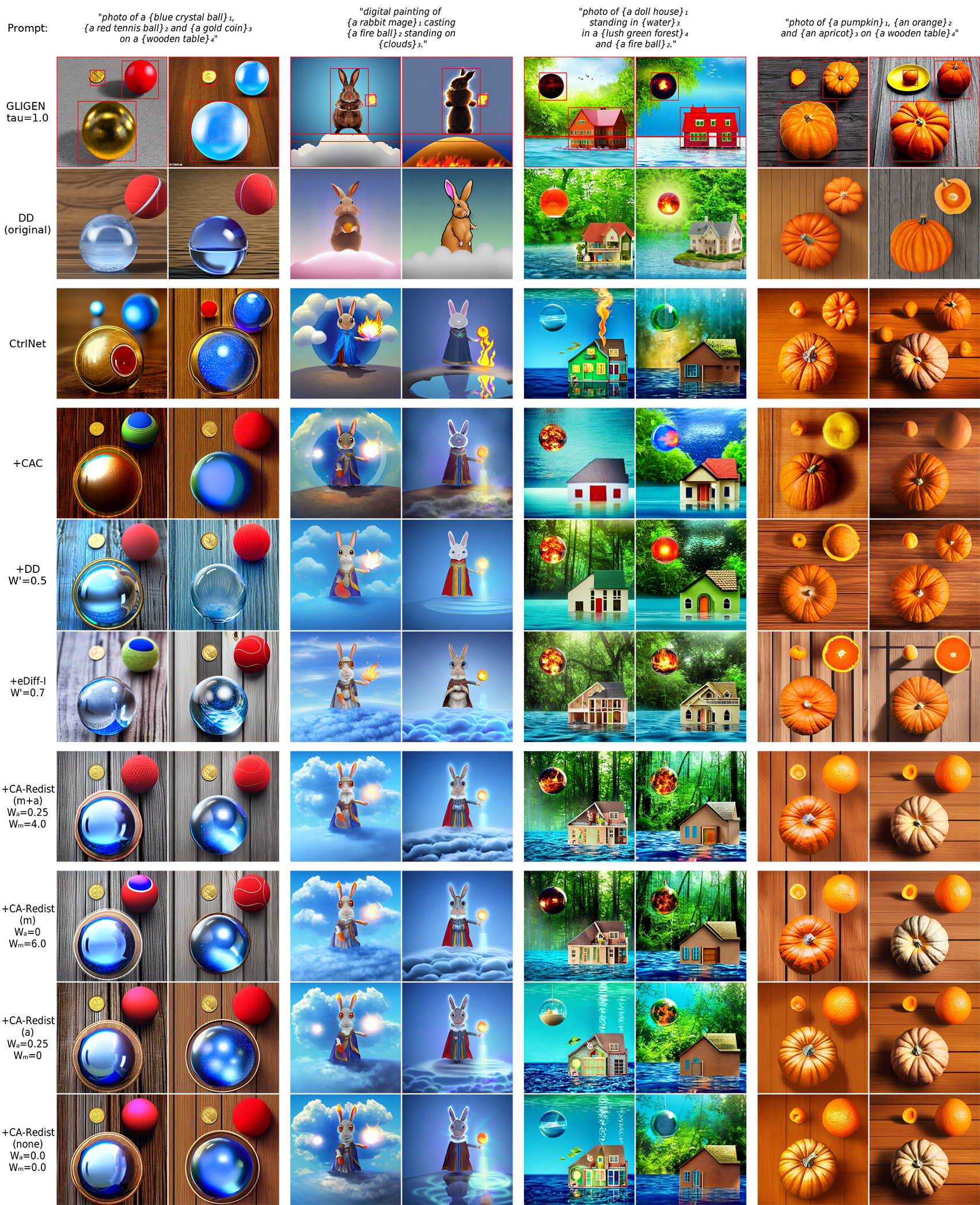}
    \caption{Qualitative comparison with ControlNet trained for scribbles.}
    \label{fig:qualitative_scribbles_full}
\end{figure*}

\section{Qualitative comparison with complex shapes}
\label{sec:supp:complex}

A comparison of different methods using more complex shapes is shown in Fig.~\ref{fig:complexfull}.
The layouts used for this comparison are given in Fig.~\ref{fig:complexlayouts}.
Note that baselines (+CAC, +DD, +eDiff-I) fail to always assign the correct descriptions to the right locations and properly separate features. 
For example, for cats and dogs, +DD generates two cats and eDiff-I, while largely performing quite well still assigns a mixture of cat- and dog-like features to the region annotated as a ``grey dog''.

\begin{figure}
    \centering
    \includegraphics[width=0.75\linewidth]{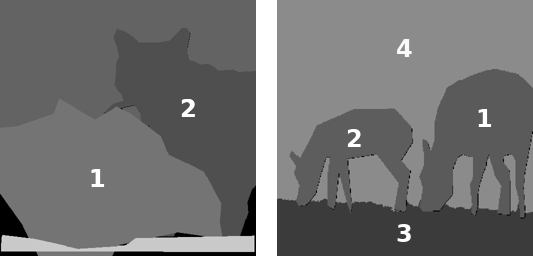}
    \caption{Layouts used in the qualitative comparison with complex shapes shown in Fig.~\ref{fig:complexfull}.
    }
    \label{fig:complexlayouts}
\end{figure}

\begin{figure*}
    \centering
    \includegraphics[width=0.99\linewidth]{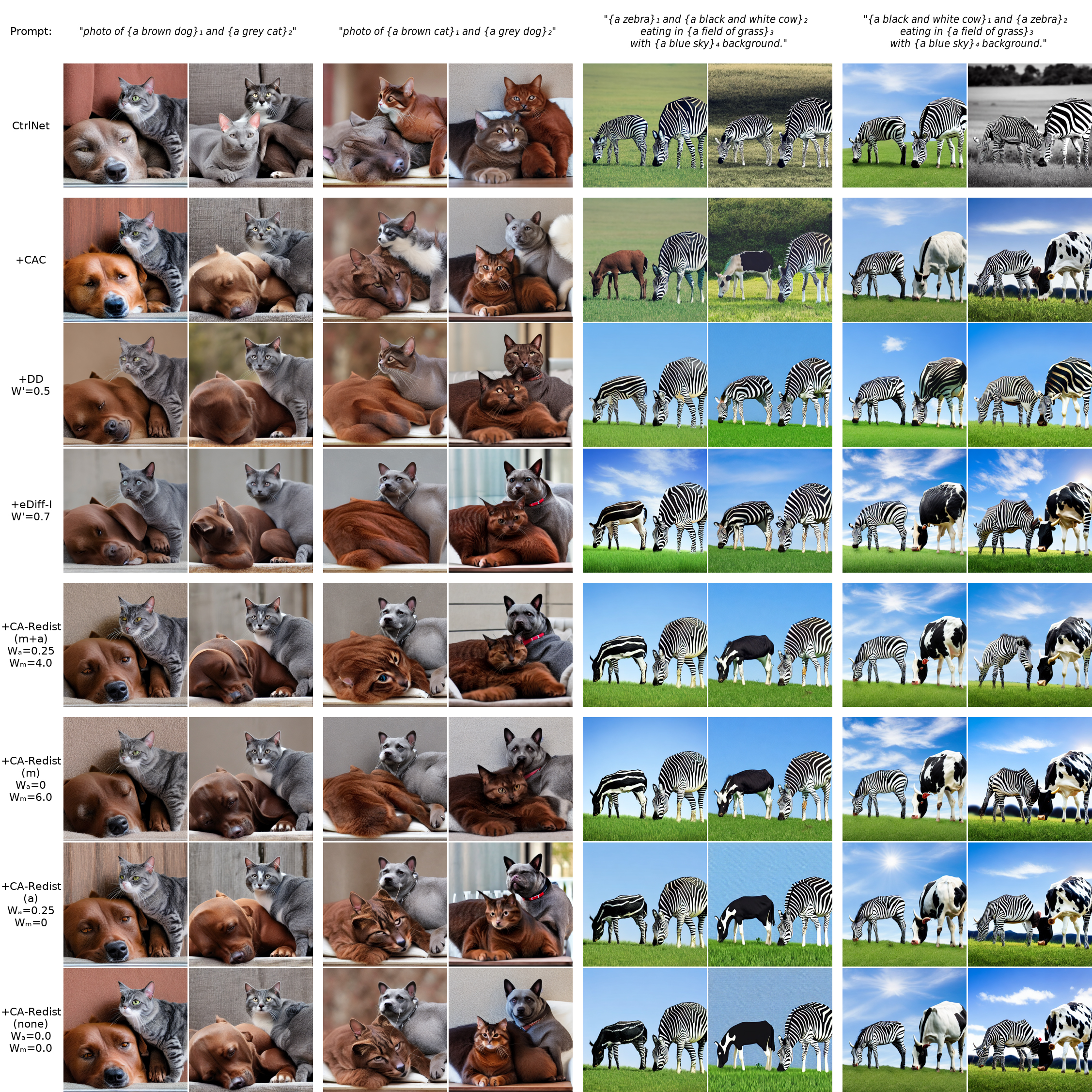}
    \caption{A qualitative comparison with complex shapes. 
    In the example for cats and dogs, region 1 corresponds to the bottom left region and region 2 to the region on the right.
    In the example for cows and zebras, region 1 corresponds to the animal shape on the right and region 2 to the animal shape on the left.
    For exact layout specifications, see Fig.~\ref{fig:complexlayouts}.
    We can observe some concept bleeding for different methods, especially evident in the cows and zebras examples, even for CA-Redist.
    Concept bleeding is more severe for the DenseDiffusion and eDiff-I baselines while the CAC baseline generated a brown animal instead of a black and white cow.
    }
    \label{fig:complexfull}
\end{figure*}

\section{Additional examples}
\label{sec:supp:more}
In ~\cref{fig:extra:1,fig:extra:2,fig:extra:3,fig:extra:4}, we provide some additional comparisons with more examples. In the prompts for these figures, we indicate the localized object phrases with curly brackets, for example ``\{a grape:TOP\_LEFT\}'', where the part after the colon specifies the position in the layout (and is included here only to clarify the prompt to the reader).
Note that the bounding boxes for GLIGEN are not part of the generated images and are included to inform the reader about the difference in input.
%Note that we chose to include the bounding boxes for GLIGEN in the final image to illustrate the different layout conditioning from the rest of the methods in the comparison.

\begin{figure*}[!htb]
    \centering
    \includegraphics[width=0.9\textwidth]{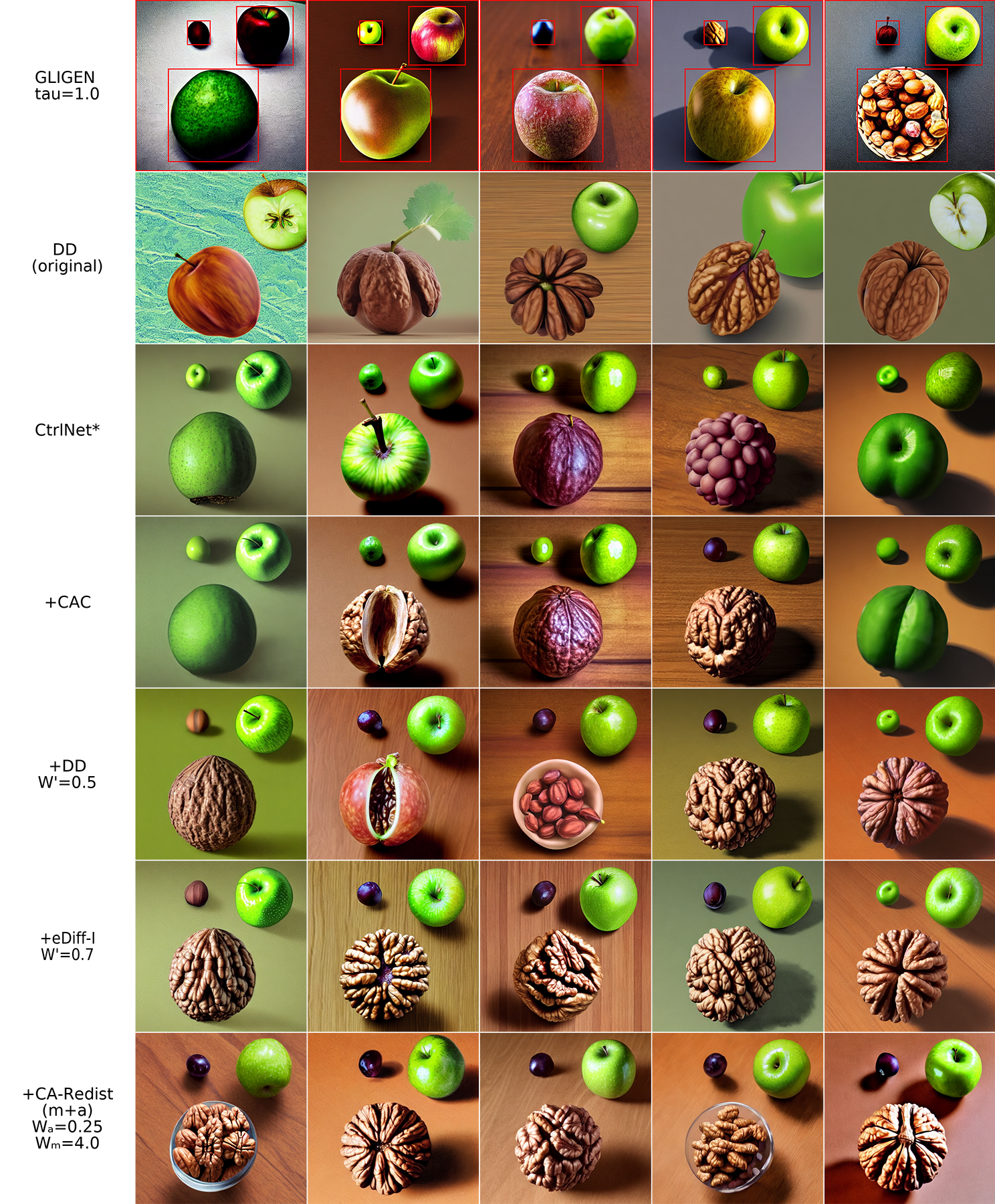}
    \caption{Comparison for prompt: ``a highly detailed photorealistic image of \{a grape:TOP\_LEFT\}, \{a walnut:BOTTOM\} and \{a green apple:TOP\_RIGHT\} on \{a kitchen table\}.''. 
    All five seeds are given to show the consistency of results.
    The results are consistent with our observations in \cref{sec:qualitative}. 
    GLIGEN and original DenseDiffusion fail to properly respect the object description, while DenseDiffusion misses out the smaller object.
    Plain ControlNet and its CAC-extended version also fail to respect object descriptions due to shape ambiguity.
    Adding DenseDiffusion-like or eDiff-I-like attention control to ControlNet yields significantly better results, but for some seeds, the assignments fail.
    CA-Redist on the other hand, generates images that are completely faithful to the localized prompt.
    See also \cref{sec:supp:more}.
    }
    \label{fig:extra:1}
\end{figure*}

% \noindent%
% \begin{minipage}{2\linewidth}% to keep image and caption on one page
% \makebox[\linewidth]{%        to center the image
%     \includegraphics[width=0.9\textwidth]{images/walnut_apple_grape.png}}
% \captionof{figure}{Comparison for prompt: ``a highly detailed photorealistic image of \{a grape:TOP\_LEFT\}, \{a walnut:BOTTOM\} and \{a green apple:TOP\_RIGHT\} on \{a kitchen table\}.''. 
%     All five seeds are given to show the consistency of results.}\label{fig:extra:1}
% \end{minipage}

\begin{figure*}[!htb]
    \centering
    \includegraphics[width=0.9\textwidth]{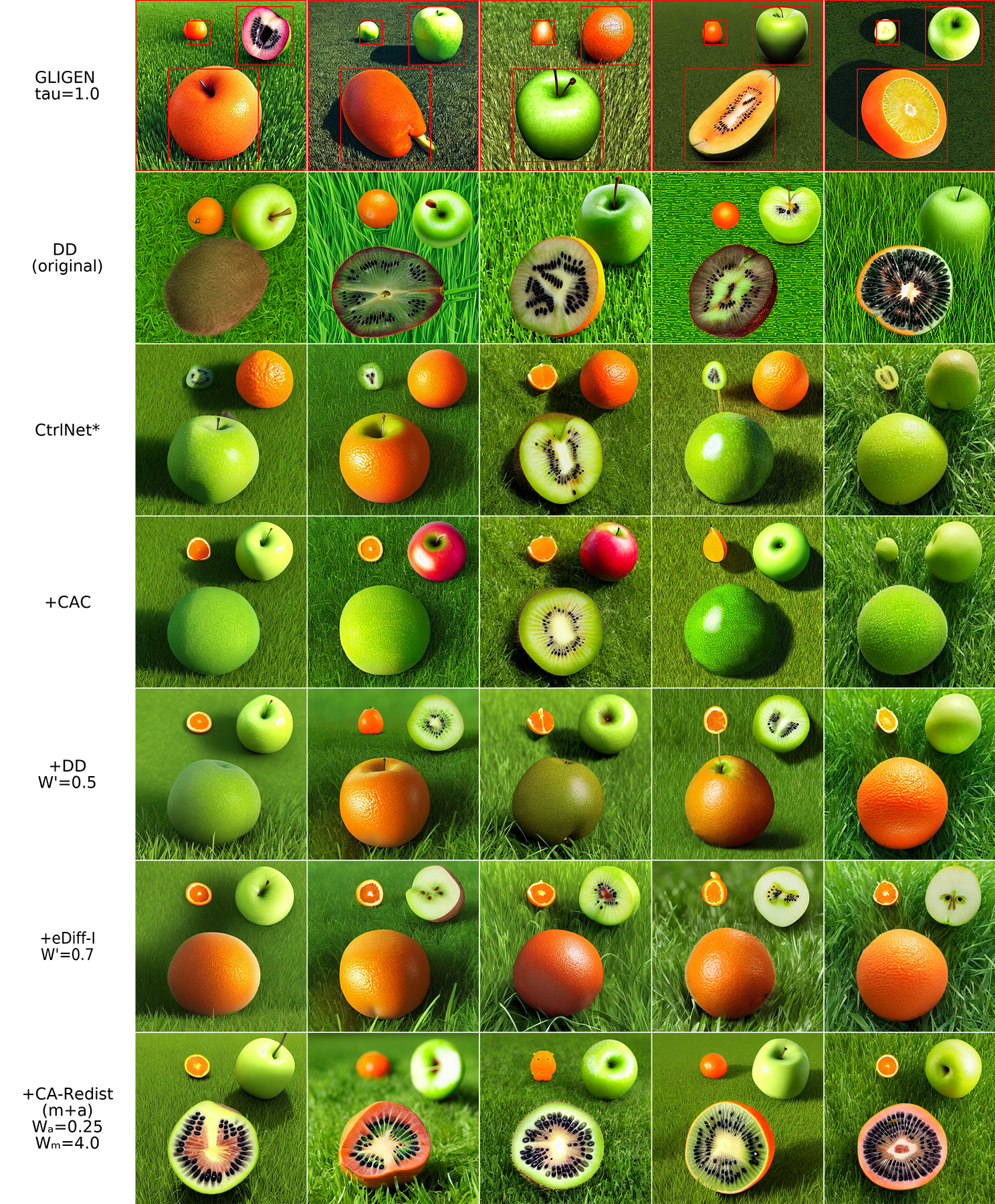}
    \caption{Comparison for prompt: ``a highly detailed photorealistic image of \{an orange:TOP\_LEFT\}, \{a kiwi:BOTTOM\} and \{a green apple:TOP\_RIGHT\} on \{grass\}.''.
    All five seeds are given to show the consistency of results.
    Here, the same results as before can be observed. Only CA-Redist is able to correctly assign the finest object details, with the exception of the third seed, where an orange figure instead of the fruit has been generated and some orange color ``leaked'' to the kiwi in the foreground for some seeds.
    In comparison, other methods largely failed to generate a kiwi in the correct place for most of the seeds (except the original Dense Diffusion).
    See also \cref{sec:supp:more}.
    }
    \label{fig:extra:2}
\end{figure*}

\begin{figure*}[!htb]
    \centering
    \includegraphics[width=0.9\textwidth]{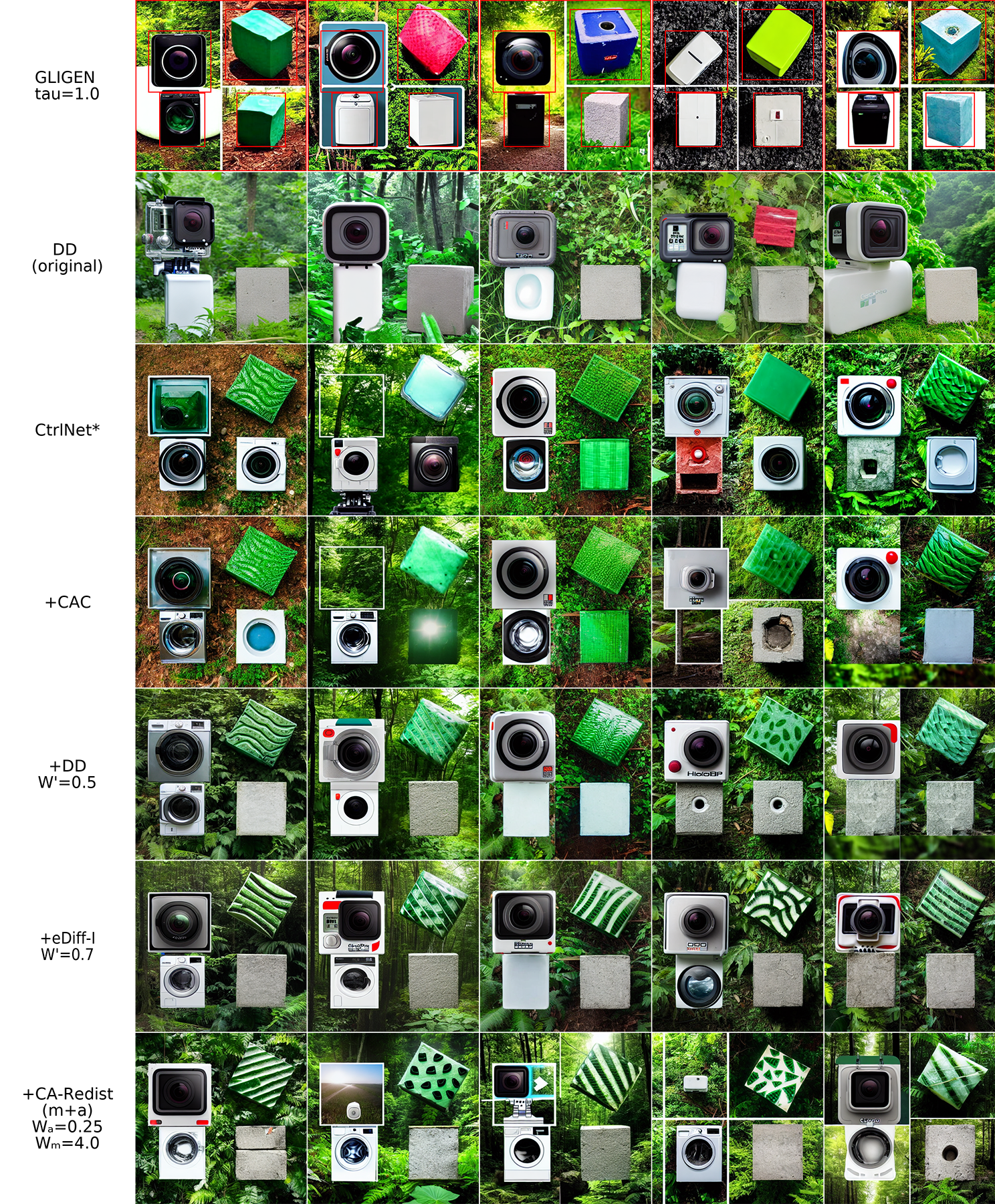}
    \caption{Comparison for prompt: ``a photo of \{a gopro camera:TOP\_LEFT\} on top of \{a washing machine:BOTTOM\_LEFT\}, and to the right \{a square watermelon:TOP\_RIGHT\} falling on \{a concrete block:BOTTOM\_RIGHT\}. Background is \{a lush forest\}.''.
    All five seeds are given to show the consistency of results.
    See also \cref{sec:supp:more}.
    }
    \label{fig:extra:3}
\end{figure*}

\begin{figure*}[!htb]
    \centering
    \includegraphics[width=0.9\textwidth]{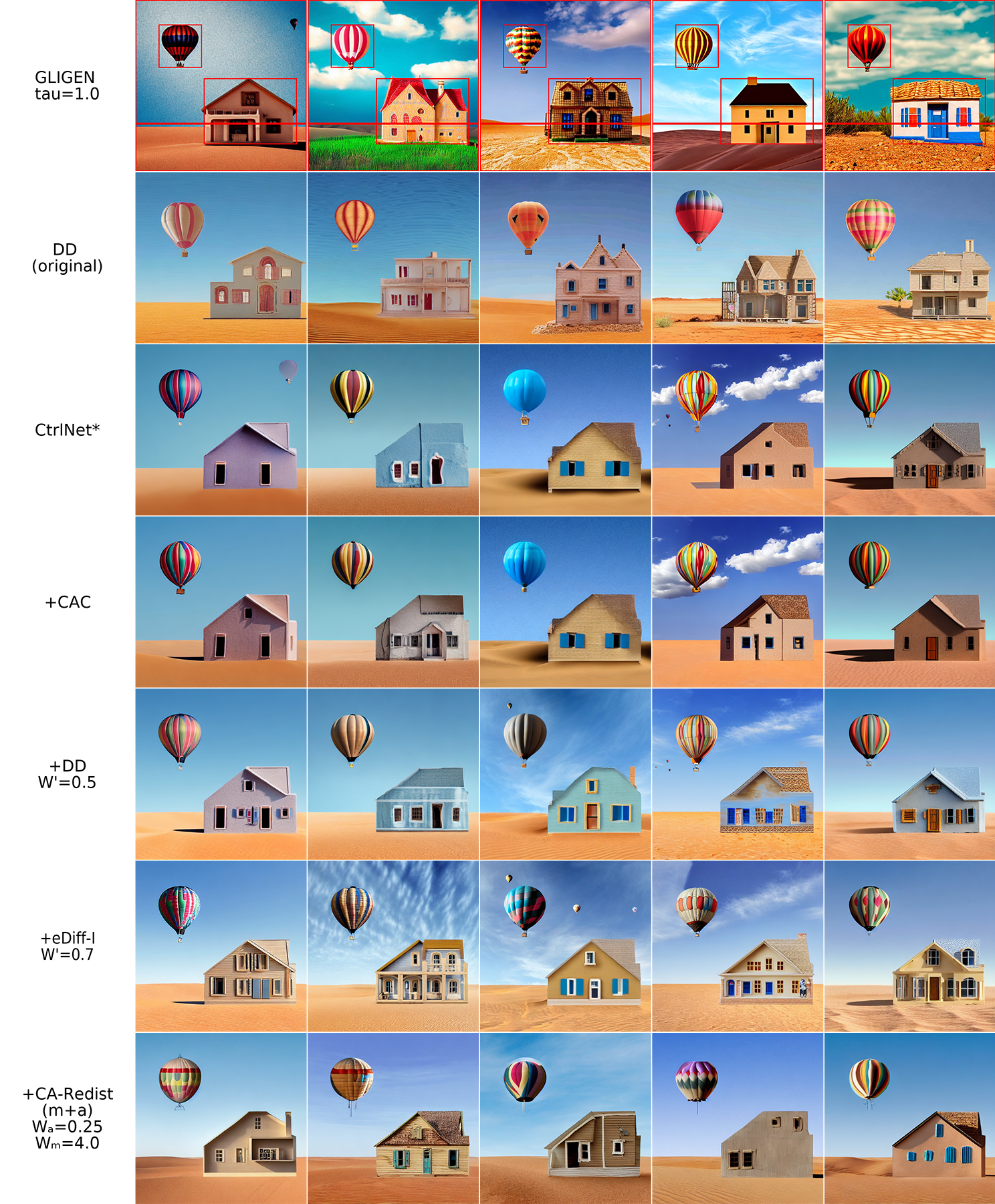}
    \caption{Comparison for prompt: ``a photo of \{a doll house:BOTTOM\_RIGHT\} \{in the desert:BOTTOM\} Background is \{a blue sky:TOP\} with \{a hot air balloon:TOP\_LEFT\}.''
    All five seeds are given to show the consistency of results.
    Here, almost all tested methods appear to work well, with the exception of GLIGEN, which does not generate a desert for some seeds.
    See also \cref{sec:supp:more}.
    }
    \label{fig:extra:4}
\end{figure*}

\end{document}